\def\year{2022}\relax
\documentclass[letterpaper]{article} 
\usepackage{aaai22}  
\usepackage{times}  
\usepackage{helvet}  
\usepackage{courier}  
\usepackage[hyphens]{url} 
\usepackage{graphicx} 
\usepackage{caption} 

\DeclareCaptionStyle{ruled}{labelfont=normalfont,labelsep=colon,strut=off} 
\usepackage{algorithm}
\usepackage{algorithmic}
\usepackage{paralist}

\usepackage{amsmath,bm}
\usepackage{multirow}
\usepackage{amssymb}
\usepackage{booktabs}
\usepackage{nameref}
\usepackage{setspace}
\usepackage{hyperref}
\usepackage{hyperref}
\usepackage{fancyhdr}
\usepackage{newfloat}
\usepackage{listings}
\floatstyle{ruled}
\newfloat{listing}{tb}{lst}{}
\floatname{listing}{Listing}

\title{Weakly Supervised Reasoning by Neuro-Symbolic Approaches}
\author {
    Xianggen Liu$^1$\quad Zhengdong Lu$^2$\quad Lili Mou$^3$
}
\affiliations {
$^1$College of Computer Science, Sichuan University, \url{liuxianggen@scu.edu.cn}\\
$^2$DeeplyCurious.ai, \url{luz@deeplycurious.ai}\\
$^3$Department of Computing Science \& Alberta Machine Intelligence Institute (Amii)\\
University of Alberta, Canada CIFAR AI Chair, \url{doublepower.mou@gmail.com}
}
\begin{document}

\maketitle

\begin{abstract}
Deep learning has largely improved the performance of various natural language processing (NLP) tasks. However, most deep learning models are black-box machinery, and lack explicit interpretation. In this chapter, we will introduce our recent progress on neuro-symbolic approaches to NLP, which combines different schools of AI, namely, symbolism and connectionism. Generally, we will design a neural system with symbolic latent structures for an NLP task, and apply reinforcement learning or its relaxation to perform weakly supervised reasoning in the downstream task. Our framework has been successfully applied to various tasks, including table query reasoning, syntactic structure reasoning, information extraction reasoning, and rule reasoning. For each application, we will introduce the background, our approach, and experimental results.

\end{abstract}

\newcommand{\newcite}[1]{\citeauthor{#1}~(\citeyear{#1})}

\renewcommand{\normalsize}{\fontsize{11pt}{0}\selectfont} 
\singlespacing

\thispagestyle{fancy}
\renewcommand{\headrulewidth}{0pt}
\rhead{}
\cfoot{\textit{Compendium of Neurosymbolic Artificial Intelligence}, 665--692, 2023. IOS Press. \url{https://doi.org/10.3233/FAIA230162}
}

\section{Introduction}~\label{sec:Intro}
\textit{Symbolism} and \textit{connectionism} are two schools of artificial intelligence (AI)~\cite{AImagazine,AImagazine22}. Symbolic AI aims at building symbolic representations and performing symbolic manipulations of  knowledge in a human-understandable manner for intelligent systems, whereas connectionists believe that intelligence lies in the connections of neurons. Therefore, connectionists build artificial neural networks, which, albeit over-simplified, mimic human brains to achieve a certain level of intelligence. 

In the past decades, connectionism appears to be the obvious winner in terms of the performance of a certain task~\cite{DL}. In machine reading comprehension, for example, the performance of deep neural networks is claimed to exceed the human level~\cite{human}. This is because deep neural networks are powerful learning machines; with multiple non-linear transformations, they can learn highly complicated features suited to the task of interest. On the contrary, symbolic AI systems would build logical representations and perform logical reasoning for the task~\cite{logic}. Such an approach typically fails to achieve satisfactory performance, because a powerful logical system (e.g., first-order logic) is undecidable~\cite{logic-textbook}, whereas a restricted system (e.g., description logic~\cite{descriptionlogic}) may be decidable but still the computational complexity makes reasoning intractable. More importantly, logical forms do not exist naturally, and the mapping from the problem domain to the logical domain is non-trivial and prone to errors. Early approaches use manually designed rules and heuristics~\cite{fuzzylogic,ExpertSystem,rule}, and recently, such a key step towards symbolism is ironically accomplished by connectionists' neural models~\cite{SP}. 

However, criticisms have been made against connectionists' neural networks, mainly due to the lack of interpretability and explainability~\cite{AImagazine22}. Seo et al.~\cite{attnMC}, for example, build excessive attention mechanisms for machine comprehension. Although some researchers demonstrate certain correlation information in attention probabilities~\cite{attn1,attn2}, others claim attention is not an interpretation~\cite{attentionNOT}. We argue that, even in the optimistic case, the classic attention in sequence-to-sequence models~\cite{jointly} fails to provide symbolic explanations beyond correlation, and thus is still black-box machinery. 

In this chapter, we present our recent development of a neuro-symbolic framework, with a focus on reasoning in natural language processing (NLP) tasks. Here, we treat \textit{reasoning} as providing intermediate thinking steps for drawing a conclusion. More importantly, such reasoning should be performed in a \textit{weakly supervised} manner, that is, the training signals only exist for the final conclusion, whereas the intermediate thinking steps are not directly supervised and have to be reasoned by the model itself. We argue that weak supervision is crucial to meaningful, non-trivial reasoning models, or otherwise, the system degrades to multi-task learning and again is black-box machinery~\cite{SP}.
Take the table query as an example~\cite{coupling}, where the input is a table and a natural language query and the output is some result obtained by SQL-like execution. The training set comprises input-output pairs, but does not have execution commands. In fact, execution commands are important intermediate thinking steps for drawing the conclusion, but are extremely difficult to annotate. Our formulation of weakly supervised reasoning enables us to reduce human labor, while achieving high interpretability of deep learning models.

\begin{figure}[!t]
\centering\includegraphics[width=0.6\linewidth]{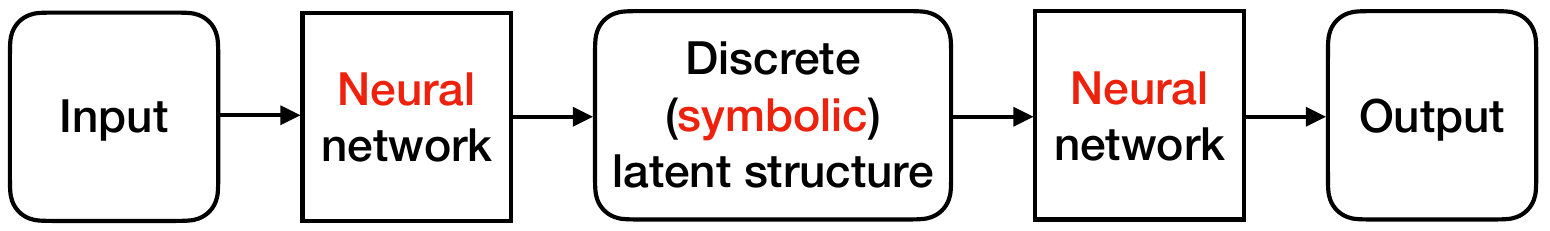}
\caption{Overview of our neuro-symbolic reasoning framework.}\label{fig:overview}
\end{figure}

Figure~\ref{fig:overview} shows the general framework of our neuro-symbolic approaches. As seen, we model reasoning (i.e., intermediate thinking steps) as discrete, symbolic latent variables. They are predicted by a neural network based on some input. Then, another neural network will predict the output based on the symbolic latent variables. The key design principles include: 1) the latent symbolic variables should play a key role in the ultimate task, and 2) there is no bypassing connection from input to output. In this way, the model has to perform meaningful intermediate reasoning to accomplish the task, and by optimizing the ultimate-task performance, we will be able to obtain meaningful latent structures as the reasoning outcome.

Training such a neuro-symbolic model is more sophisticated than back-propagation, because the discrete latent variables are not differentiable and do not propagate gradient. Reinforcement learning (RL) is a common technique for learning with discrete variables in a trial-and-error fashion~\cite{rl}. However, RL is known to have high variance, especially at the beginning of training. We thus propose neural relaxations~\cite{coupling} or pretraining methods~\cite{imitation,chunk,ruler} to obtain meaningful initialization for RL. 

Generally, our approach combines the two schools of AI, namely, symbolism and connectionism, as we
leverage connectionists' neural networks to achieve high performance, but we also let the neural network to perform symbolic reasoning in a weakly supervised manner, so that our model is interpretable. In the rest of this chapter, we will apply our neuro-symbolic framework to various applications: table query reasoning~(Section~\ref{LLM:sec:sql}), syntactic structure reasoning~(Section~\ref{LLM:sec:syntax}), information extraction reasoning~(Section~\ref{LLM:sec:IE}), and rule reasoning~(Section~\ref{LLM:sec:rule}).

\section{Table Query Reasoning}\label{LLM:sec:sql}

\begin{figure}[!t]
\includegraphics[width=\linewidth]{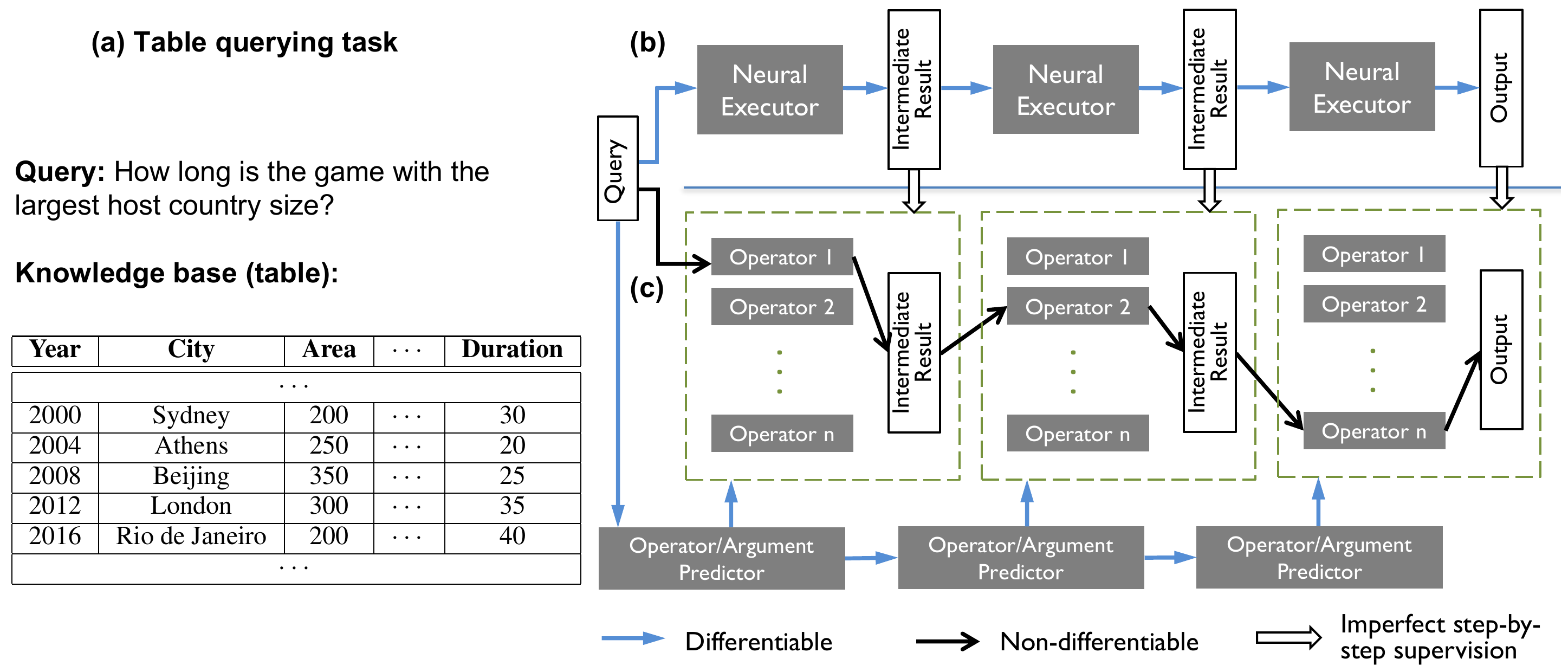}
\caption{(a) The task of table querying. (b) Neural executor. (c) Symbolic executor. }\label{fig:NEdiagram}
\end{figure}

Using natural language to query a knowledge base is an important task in NLP and has wide applications in question answering, human-computer conversation, etc.\footnote{Part of this chapter was published in~\cite{coupling}, available at \url{http://proceedings.mlr.press/v70/mou17a/mou17a.pdf}, licensed under \href{https://creativecommons.org/licenses/by/4.0/}{CC BY 4.0}. Changes are made to fit this book chapter.} Figure~\ref{fig:NEdiagram}a illustrates an example of a knowledge base (a table) and a query ``How long is the game with the largest host country size?'' To answer the question, we should first find a row with the largest value in the column \textit{Area}, and then select the value of the chosen row with the column being \textit{Duration}.

A typical approach to table querying, known as \textit{semantic parsing}, is to convert a natural language sentence to an executable logic form, which could be thought of intermediate thinking steps to draw the conclusion. Traditionally, building a semantic parser requires extensive human engineering of explicit features~\cite{FB}. With the fast development of deep learning, an increasing number of studies use neural networks for semantic parsing. Dong et al.~\cite{SP} apply sequence-to-sequence (seq2seq) neural models to generate a logic form conditioned on an input sentence, but the training requires groundtruth logic forms, which are costly to obtain and speciﬁc to a certain dataset. In a realistic setting, we only assume groundtruth denotations (i.e., execution results) are available, and that we do not know execution sequences or intermediate execution results. Liang et al.~\cite{liang2016neural} train a seq2seq network by REINFORCE policy gradient. But it is known that the REINFORCE algorithm is sensitive to the initial policy; also, it could be very difﬁcult to get started at early stages.

Yin et al.~\cite{NE} propose a fully distributed neural enquirer, comprising several neuralized execution layers of field attention, row annotation, etc. The model can be trained in an end-to-end fashion because all components are differentiable. However, it lacks explicit interpretation and is not efficient in execution due to intensive matrix/vector operation during neural processing. Neelakantan et al.~\cite{NP} propose a neural programmer by defining a set of symbolic operators (e.g., \texttt{argmax}, \texttt{greater\_than}); at each step, all possible execution results are fused by a softmax layer, which predicts the probability of each operator at the current step. The step-by-step fusion is accomplished by weighted average and the model is trained with mean square error. Hence, such approaches work with numeric tables, but may not be suited for other operations like string matching. It also suffers from the problem of ``exponentially growing combinatorial states,'' as the model explores the entire space at a time by step-by-step weighted average. 

In this chapter, we propose a neuro-symbolic approach to weakly supervised semantic parsing. This follows our general framework (Figure~\ref{fig:overview}), as the execution steps are treated as discrete/symbolic latent structure. In addition, we propose to couple neuralized and symbolic execution for table querying, where symbolic execution\footnote{The sense of \textit{symbolic execution} here should not be confused with ``symbolic execution of a program'' (see \url{https://en.wikipedia.org/wiki/Symbolic_execution} for example).} defines symbolic operators and keeps discrete intermediate execution results, whereas neuralized/distribution execution mimics the process by neural networks. Our intuition arises from the observation that a fully distributed/neuralized executor also exhibits some (imperfect) symbolic interpretation. For example, the field attention gadget in \cite{NE} generally aligns with column selection.
We therefore use the neural executor's intermediate execution results as supervision signals to pretrain a symbolic executor. Guided by such imperfect step-by-step supervision, the symbolic executor learns a fairly meaningful initial policy, which largely alleviates the cold start problem of REINFORCE.

\subsection{The Proposed Neuro-Symbolic Execution}
In this part, we will introduce our approach in detail, including the fully distributed neural executor, the symbolic executor, and their training process.

\begin{figure}[!t]\centering
\includegraphics[width=.8\linewidth]{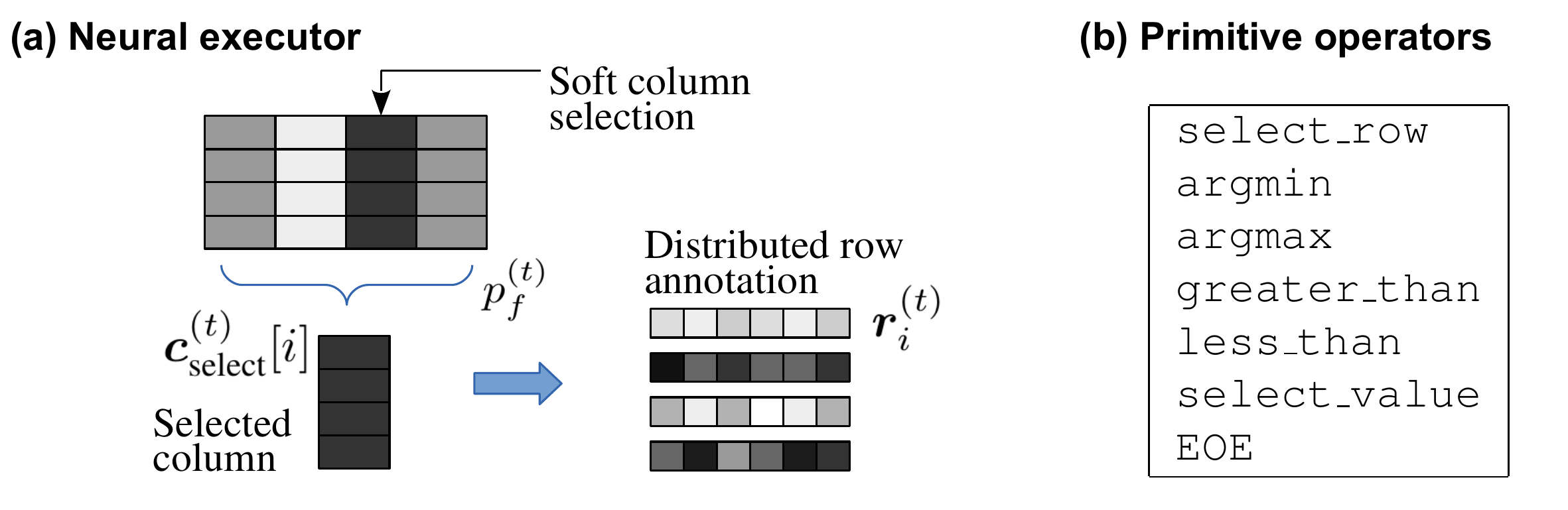}
\caption{(a) The details of neural execution. (b) Primitive operators of a symbolic executor. }\label{fig:NEdiagram2}
\end{figure}

The \textbf{distributed executor} makes full use of neural networks for table querying.
By ``distributed,'' we mean that all semantic units (including words in the query, entries in the table, and execution results) are represented as distributed, real-valued vectors and processed by neural networks.
One of the most notable studies of distributed semantics is word embeddings, which map discrete words to vectors as meaning representations~\cite{mikolov2013distributed}.
The distributed executor consists of the following main components.
\begin{compactitem}
	\item Query encoder. Words are mapped to word embeddings and a bi-directional recurrent neural network (RNN) aggregates information over the sentence. RNNs' last states in both directions are concatenated as the query representation $\bm q$.
	\item Table encoder. All table cells are also represented as embeddings. For a cell $c$ (e.g., \textit{Beijing} in Figure~\ref{fig:NEdiagram}) with its column/field name being $f$ (e.g., \textit{City}), the cell vector is the concatenation of the embeddings of $c$ and $f$, further processed by a multi-layer perceptron. We denote the representation of a cell as $\bm c$.
	\item Executor.  This part comprises several steps of distributed execution, shown in Figure~\ref{fig:NEdiagram2}a. In each execution step $t$, the neural network selects a column by softmax attention $p_{f_j}^{(t)}$ for some field $f_j$ and annotates the a row as $\bm r_i^{(t)}$ for some row $i$.
\begin{align}
p_{f_j}^{(t)}&=\operatorname{softmax}\left(\operatorname{MLP}\big([\bm q; \bm f_j; \bm g^{(t-1)}]\big)\right)=\frac{\exp\{
	\operatorname{MLP}([\bm q; \bm f_j; \bm g^{(t-1)}])
	\}
}{   \sum_{j'} \exp\left\{
\operatorname{MLP}\left([\bm q; \bm f_{j'}; \bm g^{(t-1)}]\right)
\right\}
}
\label{eqn:fieldselection}
\end{align}
where $\bm f_j$ is the embedding of the field name $f_j$, and $\bm g^{(t-1)} = \operatorname{MaxPool}_i \{\bm r_i^{(t-1)}\}$ is the global information of the previous step. The softly selected column can be represented as $\bm c_\text{select}^{(t)}[i]=\sum_j p_{f_j}^{(t)}\bm c_{ij}$.
	Then, the neural executor annotates each row with a real-valued vector $\bm r_i^{(t)}=\operatorname{MLP}([\bm q; \bm g^{(t-1)}; \bm r_i^{(t-1)}; \bm c_\text{select}^{(t)}[i]])$. The row vector can be intuitively thought of as row selection in query execution, but is represented by distributed semantics in our model.\footnote{In a pilot experiment, we tried a gating mechanism to indicate the results of row selection in hopes of aligning symbolic table execution. However, preliminary results show that such gates do not exhibit much interpretation, but lead to performance degradation. The distributed semantics provide more information than a 1-bit gate for a row.
	}
	In the last step of execution, a softmax classifier is applied to the entire table to select a cell as the answer. 
\end{compactitem}

\medskip

Our \textbf{symbolic executor} is defined with a set of primitive operators for the task (Figure~\ref{fig:NEdiagram2}b); it uses a machine learning model to predict the operator sequence and its arguments.
Our symbolic executor is different from the neural programmer \cite{NP} in that we keep symbolic execution results, whereas the neural programmer fuses execution results by weighted average.

Stacked with multiple primitive operators, the executor can answer fairly complicated questions like ``How long is the last game which has smaller country size than the game whose host country GDP is 250?'' In this example, the execution sequence is: 1) \texttt{select\_row}: select the row where the column is \textit{GDP} and the value is mentioned in the query, 2) \texttt{less\_than}: select rows whose country size is less than that of the previously selected row, 3) \texttt{argmax}: select the row whose year is the largest among previously selected rows, and 4) \texttt{select\_value}: choose the value of the previously selected row with the column being \textit{Duration}. Then the execution terminates. In our experiment, the execution is limited to four steps (\texttt{EOE} excluded) as such queries are already complicated in terms of logical depth.

The symbolic execution is controlled by a neural network. In particular, we use RNNs to predict the operator and the field 
\begin{align}
\bm h_\text{op}^{(t)} &= \operatorname{sigmoid}(W_\text{op}^\text{(rec)}\bm h_\text{op}^{(t-1)}), \quad p_{\text{op}_i}^{(t)} = \operatorname{softmax}\left\{\bm w_{\text{op}_i}^\text{(out)}{}^\top\bm h_\text{op}^{(t)}\right\}\label{eqn:op_softmax}\\
\bm h_\text{field}^{(t)} &= \operatorname{sigmoid}(W_\text{field}^\text{(rec)}\bm h_\text{field}^{(t-1)}), \quad
p_{f_j}^{(t)} = \operatorname{softmax}\{\bm  f_j^\top\bm h_\text{field}^{(t)}\}\label{eqn:fd_softmax}
\end{align}
where $\bm h$s are hidden states. 

Training a symbolic executor without step-by-step supervision signals is non-trivial.
A typical training method is reinforcement learning in a trial-and-error fashion. However, for a random initial policy, the probability of recovering an accurate execution sequence is extremely low. Given a $10\times10$ table, for example, the probability is $1/(6^4\cdot 10^4)\approx7.7\times10^{-8}$; the probability of obtaining an accurate denotation is 1\%, which is also very low. Therefore, symbolic executors are difficult to train.

\smallskip
\textbf{A unified view.}
We now have two worlds of execution: 1) The distributed executor is end-to-end learnable, but it is of low execution efficiency  because of intensive matrix/vector multiplication during neural information processing. It also lacks explicit interpretation. 2) The symbolic executor has high execution efficiency and explicit interpretation. However, it  cannot be trained in an end-to-end manner, and suffers from the cold start problem of reinforcement learning.
We propose to combine the two worlds by using the distributed executor's intermediate execution results to pretrain the symbolic executor for an initial policy. Then, REINFORCE can be used for policy improvement. 

Our observation is that the field attention in Eqn.~(\ref{eqn:fieldselection}) generally aligns with column selection in Eqn.~(\ref{eqn:fd_softmax}). We therefore pretrain the column selector in the symbolic executor with labels predicted by a fully distributed executor (Figure~\ref{fig:NEdiagram}a). This is given by the cross-entropy loss $J=-\sum_{i=1}^m\sum_{j=1}^{n_{\text{label}}^{(i)}} \hat{t}_j^{(i)} \log p_j^{(i)}\label{eqn:fieldloss}
$,
where $n_{\text{label}}^{(j)}$ is the number of labels (all columns) for the $j$th step. $m$ stands for the number of actions and
$\bm p^{(i)}\in\mathbb{R}^{n_{\text{label}}^{(i)}}$ is the predicted probability by the symbolic predictor (Figure~\ref{fig:NEdiagram}c). In our scenario, we only pretrain column predictors.

After obtaining a meaningful, albeit imperfect, initial policy, we apply REINFORCE~\cite{williams92-reinforce} to improve the policy. We define a binary reward $R$ indicating whether the final result of symbolic execution matches the groundtruth denotation. The loss function of a policy is the negative expected reward, where actions $a_1, a_2, \cdots, a_n$---including operators~(\ref{eqn:op_softmax}) and fields~(\ref{eqn:fd_softmax})---are sampled from the current predicted probabilities $J=-\mathbb{E}_{a_1, a_2, \cdots, a_n}[R(a_1, a_2, \cdots, a_n)]$.
In this way, we can effectively train an interpretable symbolic executor.

\subsection{Experimental Results}
	We evaluated our approach on a synthetic QA dataset in~\cite{NE}. The dataset comprises 25K different tables and queries for training; validation and test sets contain 10K samples, respectively, and do not overlap with the training data. Each table is of size $10\times 10$, but different samples have different tables; the queries can be divided into four types: \texttt{SelectWhere}, \texttt{Superlative}, \texttt{WhereSuperlative}, and \texttt{NestQuery}, requiring 2--4 execution steps (\texttt{EOE} excluded). 
	
\begin{table*}[!t]
	\centering
	\resizebox{\linewidth}{!}{
		\begin{tabular}{l|c|ccc|ccc}
			\toprule
			&              &\multicolumn{3}{c|}{\textbf{Denotation}} &\multicolumn{3}{c|}{\textbf{Execution}}\\
			\textbf{Query type} & \sc Sempre$^\dag$ & \textbf{Distributed}$^\dag$ & \textbf{Symbolic}& \textbf{Coupled} & \textbf{Distributed} & \textbf{Symbolic} &\textbf{Coupled}\\
			\midrule
			\tt SelectWhere &     93.8  & 96.2   & \ \ 99.2       & \textbf{\ \ 99.6} & -- &\ \ 99.1&\textbf{\ \ 99.6} \\
			\tt Superlative &     97.8  & {98.9} & \textbf{100.0}       & \textbf{100.0}& -- & \textbf{100.0} & \textbf{100.0}\\
			\tt WhereSuperlative & 34.8 & 80.4   & 51.9       &  \textbf{\ \ 99.9} & --& \ \ \ \ 0.0 & \textbf{\ \ 91.0} \\
			\tt NestQuery     &    34.4 & 60.5  & 52.5 &  \textbf{100.0}   &  -- &\ \ \ \ 0.0&\textbf{100.0}\\
			\midrule
			Overall      &     65.2 & 84.0  & 75.8 & \textbf{\ \ 99.8} & -- &\ \ 49.5&\textbf{\ \ 97.6}\\
			\bottomrule
		\end{tabular}}			\caption{Accuracies (in percentage) of the \textsc{Sempre} tookit, the distributed neural executor, the symbolic executor, and our coupled approach. \protect$^\dag$Results reported in~\cite{NE}.}

\label{tab:SPresult}

	\end{table*}
	
	Table~\ref{tab:SPresult} presents the experimental results of our coupled approach as well as baselines.
	Because reinforcement learning is much noisier to train, we report the test accuracy corresponding to the highest validation accuracy among three different random initializations. This is also known as a \textit{restart} strategy for non-convex optimization.
	
	As we see, both distributed and symbolic executors outperform the \textsc{Sempre} system, showing that neural networks can capture query information more effectively than human-engineered features.
	Further, the coupled approach also significantly outperforms both of them. If trained solely by REINFORCE, the symbolic executor can recover the execution sequences for simple questions (\texttt{SelectWhere} and \texttt{Superlative}). However, for more complicated queries, it only learns the last one or two steps of execution and has trouble in recovering early steps. This results in low execution accuracy but near 50\% denotation accuracy because, in our scenario, we still have half a chance to obtain an accurate denotation even if the nested (early) execution is wrong---the ultimate result is either in the candidate list or not, given a wrong where-clause execution. By contrast, the coupled training largely improves the symbolic executor's performance in terms of all query types.
	
	The accuracy of execution is crucial to the interpretability of a model. We say an execution is \textit{accurate}, if all actions (operators and arguments) are correct.
	As shown above, an accurate denotation does not necessarily imply an accurate execution. We find that coupled training recovers most correct execution sequences, that the symbolic executor alone cannot recover complicated cases, and that a fully distributed model does not have explicit interpretations of execution. The results demonstrate high interpretability of our approach, which is helpful for human understanding of the execution processes.
	
\begin{figure}[!t]
	\includegraphics[width=\linewidth]{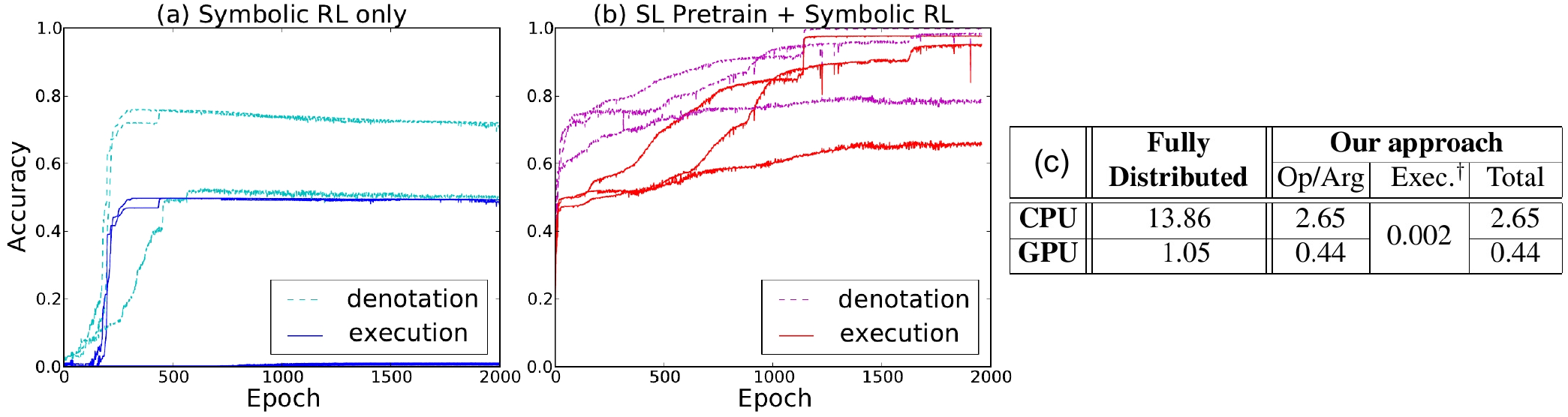}
	\caption{(a) Learning curve of reinforcement learning (RL) only. (b) RL with supervised pretraining from neuralized execution. (c) Inference efficiency. $^\dag$Accessed by C++ programming.}\label{fig:NEresult}
	\end{figure}

	We plot in Figure~\ref{fig:NEresult} the validation learning curves of the symbolic executor, trained by either reinforcement learning alone or our coupled approach. Figure~\ref{fig:NEresult}a shows that the symbolic executor is hard to train by REINFORCE alone: one trajectory obtains near-zero execution accuracy after 2000 epochs; the other two take 200 epochs to get started to escape from initial plateau. Even if they achieve $\sim$50\% execution accuracy (for simple query types) and $\sim$75\% denotation accuracy, they are stuck in the poor local optima. Figure~\ref{fig:NEresult}b presents the learning curves of the symbolic executor pretrained with intermediate field attention of the distributed executor. Since the operation predictors are not pretrained, the denotation accuracy is near 0 before reinforcement learning. However, after only a few epochs of REINFORCE training, the performance increases sharply and achieves high accuracy gradually.
	
		Figure~\ref{fig:NEresult}c compares the execution efficiency of a distributed executor and our coupled approach. All neural networks are implemented in Theano with a TITAN Black GPU and Xeon e7-4820v2 (8-core) CPU; symbolic execution is assessed in \texttt{C++} implementation. The comparison makes sense because the Theano platform is not specialized in symbolic execution, and fortunately, execution results do not affect actions in our experiment. Hence, they can be easily disentangled.	As shown in the table, the execution efficiency of our approach is 2--5 times higher than the distributed executor, depending on the implementation. The distributed executor is slow for inference, because it maps every token to a distributed real-valued vector, resulting in intensive matrix-vector operations.
	The symbolic executor only needs a neural network to predict actions (operators and arguments), and thus is more lightweight. Further, we observe the execution itself is fast, implying that, compared with distributed models, our approach could achieve even more efficiency boost with a larger table or more complicated operations.
	
	\subsection{Summary} 	In this section, we proposed a neuro-symbolic approach to weakly supervised table querying, where the training signals are provided for the execution results and we treat execution commands as discrete latent structures. We further proposed a coupled view of distributed and symbolic execution. By pretraining with intermediate execution results of a distributed executor, we manage to accelerate the symbolic model's REINFORCE training to a large extent. Our proposed approach takes advantages of both distributed and symbolic worlds, achieving high interpretability, high execution efficiency, high learning efficiency, as well as the high accuracy. 
\section{Syntactic Structure Reasoning}\label{LLM:sec:syntax}

This section concerns reasoning about syntactic structures of natural language.\footnote{Part of this chapter was published in~\cite{imitation}, available at \url{https://aclanthology.org/P19-1338/}, licensed under \href{https://creativecommons.org/licenses/by/4.0/}{CC BY 4.0}. \copyright2019 Association for Computational Linguistics. Changes are made to fit this book chapter.} From a linguistic perspective, a natural language sentence can be thought of as a set of nested constituents in the form of a tree structure \cite{partee2012mathematical}. 
When a parser is trained on labeled treebanks, the predicted constituency trees are useful for various NLP tasks, including relation extraction \cite{verga2016relation}, text simplification \cite{narayan2014hybrid}, and machine translation \cite{treemt}. However, expensive expert annotations are usually required to create treebanks.

\textit{Unsupervised parsing} (also known as \textit{grammar induction} or \textit{latent tree learning}) aims to learn syntactic structures without access to a treebank during training, with potential uses in low resource or out-of-domain scenarios.
In early approaches, unsupervised parsers were trained by optimizing the marginal likelihood of sentences~\cite{Klein:Manning:04}. 
More recent deep learning approaches \cite{rl-spinn, maillard2017jointly, choi} obtain latent tree structures by reinforcement learning (RL). 
Typically, this involves a secondary task, e.g., a language modeling objective or a semantic task.
However, researchers~\cite{isitsyntax} have pointed out that these methods do not yield linguistically plausible structures, and have low self-agreement when randomly initialized multiple times.

Recently, Shen et al.~\cite{prpn} proposed the parsing-reading-predict network (PRPN), which performs language modeling with structured attention. 
The model uses heuristics to induce tree structures from attention scores, and in a replication was found to be the first latent tree model to produce syntactically plausible structures \cite{replication}. 
Structured attention in the PRPN is formalized as differentiable continuous variables, making the model easy to train. 
But a major drawback is that the PRPN does not model tree-building operations directly. 
These operations need to be stipulated externally, in an ad hoc inference procedure which is not part of the model and cannot be trained.

In this section, we propose an imitation learning framework that combines the continuous PRPN with a Tree-LSTM model with discrete parsing actions, both trained without access to labeled parse trees. We exploit the advantages of the PRPN by transferring its knowledge to a discrete parser which explicitly models tree-building operations. We accomplish the knowledge transfer by training the discrete parser to mimic the behavior of the PRPN. Its policy is then refined using straight-through Gumbel-Softmax~(ST-Gumbel) \cite{st-gumbel} trained with a semantic objective, viz., natural language inference~(NLI). Our work follows the neuro-symbolic framework in Figure~\ref{fig:overview} as we treat the syntactic structure as the latent variable, whereas the downstream task is NLI classification. 
It also extends the idea in Section~\ref{LLM:sec:sql}, as we couple diverse models at the intermediate output level (latent trees in our case); its flexibility allows us to make use of heterogeneous models, such as the PRPN and the Tree-LSTM.

We evaluate our approach on the All Natural Language Inference dataset and show that it achieves a new state of the art in terms of parsing $F$-score, outperforming our base models, including the PRPN. Our work also shows that semantic objectives can improve unsupervised parsing, contrary to earlier claims \cite{isitsyntax,replication}.  

\subsection{Methodology}

\paragraph{Parsing-reading-predict network (PRPN).}
The first ingredient of our approach is the PRPN, which is trained using a language modeling objective, i.e., it predicts the next word in the text, based on previous words.

The PRPN introduces the concept of \textit{syntactic distance} $d_t$, defined as the height of the common ancestor of $w_{t-1}$ and $w_t$ in the tree ($t$ is the position index in a sentence $w_1, ..., w_N$). 
Since gold standard $d_t$ is not available, the PRPN learns the estimated $\widehat d_t$ end-to-end in an unsupervised manner.
The PRPN computes the differences between $\widehat d_t$ at the current step and all previous steps $\widehat d_j$ for $2 \le j < t$. The differences are normalized to $[0,1]$ and used to compute attention scores from right to left. These scores are applied to reweight another set of inner-sentence attention scores, which are then used in a recurrent neural network to predict the next word. Interested readers are referred to \cite{prpn} for the details.

Based on the real-valued syntactic distances in the PRPN, an external procedure is used to infer tree structures.
The main text of \cite{prpn} suggests using the following intuitive scheme: find the largest distance $\widehat d_i$ and split the sentence into two constituents $(\cdots, w_{i-1})$ and $(w_i, \cdots)$. This process is then repeated recursively on the two new constituents.
The trees inferred by this scheme, however, yield poor parsing $F$-scores, and the results reported by \cite{prpn} are actually obtained by a different scheme (evidenced in their supplementary material and code repository): find the largest syntactic distance $\widehat d_i$ and obtain two constituents $(\cdots, w_{i-1})$ and $(w_i,\cdots)$. 
If the latter constituent contains two or more words, then it is further split into $(w_i)$ and $(w_{i+1}, \cdots)$, regardless of the syntactic distance $\widehat d_{i+1}$. This scheme introduces a bias for right-branching trees, which presumably is the reason why it yields good parsing $F$-scores for English.

The reliance on this trick illustrates the point we make in the Introduction: syntactic distance has the advantage of being a continuous value, which can be computed as an attention score in a differentiable model. However, this comes at a price: the PRPN does not model trees or tree-building operations directly. These operations need to be stipulated externally in an ad hoc inference procedure. This procedure is not part of the model and cannot be trained, but yet is crucial for good performance.

\begin{figure}[!t]\centering
	\includegraphics[width=.6\linewidth]{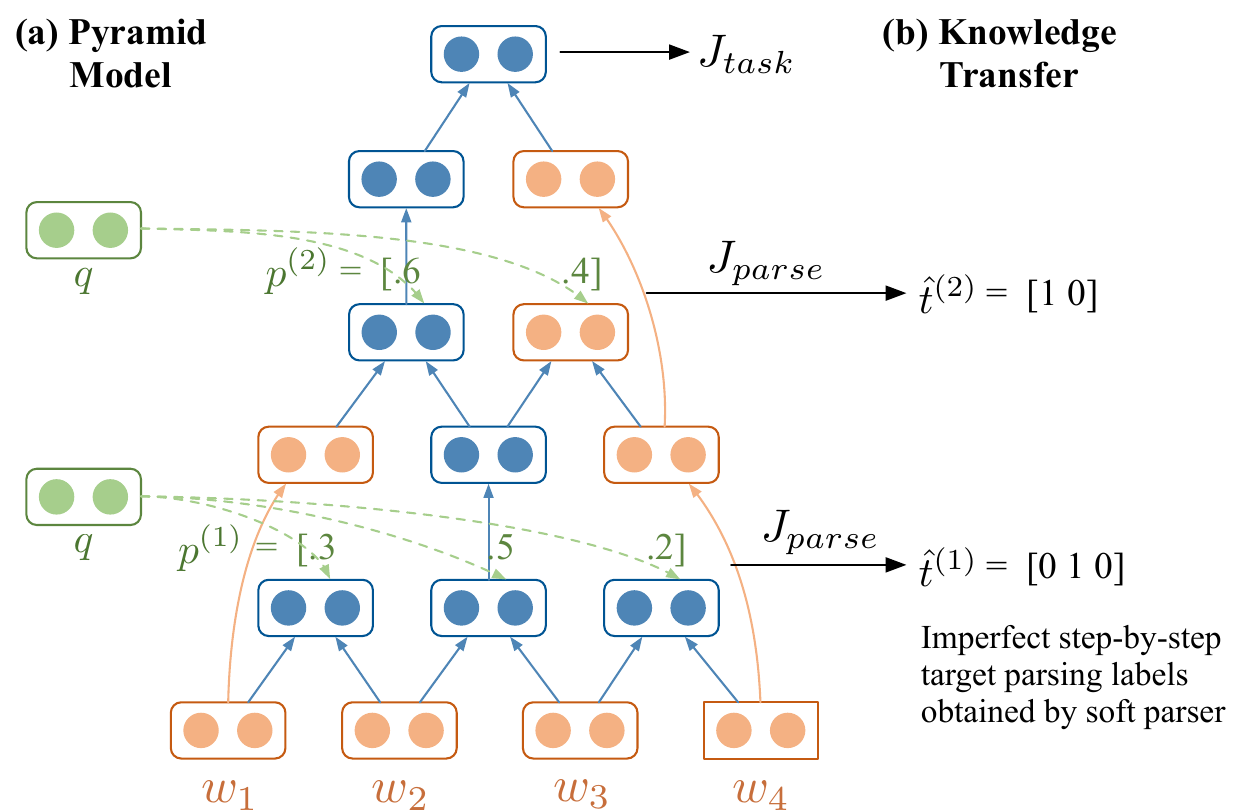}
	\caption{Overview of our approach. (a)~The Tree-LSTM model of \cite{choi}. (b)~The model is first trained with step-by-step supervision, and then Gumbel-Softmax is applied to refine the policy.}\label{fig:syntax_overview}
\end{figure}

\paragraph{Discrete syntactic parser.}
To address this problem, we combine the PRPN with a parser which explicitly models tree-building operations. Specifically, we use the pyramid-shaped, tree-based long short-term memory~(Tree-LSTM, Figure~\ref{fig:syntax_overview}a)~\cite{choi}, where reinforcement learning (RL) in this model can be relaxed by Gumbel-Softmax.

Concretely, let $\bm w_1,\bm w_2, \cdots,\bm w_N$ be the embeddings of the words in a sentence. The model tries every possible combination of two consecutive words by the Tree-LSTM, but then uses softmax (in $N-1$ ways) to predict which composition is appropriate at this step. 

Let $\bm h_1^{(1)},\cdots,\bm h_{N-1}^{(1)}$ be the candidate Tree-LSTM composition at the bottom layer. With $\bm q$ being a trainable query vector, the model computes a distribution $\bm p$:
\begin{equation}
p_i^{(1)}=\operatorname{softmax}\{\bm q^\top\bm h_i^{(1)}\}
\end{equation}
Assuming the model selects an appropriate composition at the current step, we copy all other words intact, shown as orange arrows in Figure~\ref{fig:syntax_overview}a. 
This process is applied recursively, forming the structure in the figure.

The Tree-LSTM model is learned by straight-through Gumbel-Softmax, which resembles RL as it samples actions from its predicted probabilities, exploring different regions of the latent space other than a maximum a \textit{posteriori} tree. 
Training involves doubly stochastic gradient descent: the first stochasticity comes from sampling input from the data distribution, and the second one from sampling actions for each input. 
Therefore, ST-Gumbel is difficult to train (similar to RL), and may be stuck in poor local optima, resulting in low self-agreement for multiple random initializations \cite{isitsyntax}.

\paragraph{Imitation learning.} 
Our aim is to combine the PRPN and its continuous notion of syntactic distance with a parser that has discrete tree-building operations. The mapping from the sequence of Tree-LSTM composition operations to a tree structure is not injective. Given a parse tree, we may have multiple different composition sequences, e.g., left-to-right or right-to-left. This ambiguity could confuse the Tree-LSTM during training. We solve this problem by using the PRPN's notion of syntactic distance. 

Given a parse tree predicted by the PRPN, if more than one composition is applicable, we always group the candidates with the lowest syntactic distance.
In this way, we can unambiguously determine the composition order from the trees inferred by the PRPN.
Then, we train the Tree-LSTM model in a \textit{step-by-step}~(SbS) supervised fashion.
Let $\widehat{\bm t}^{(j)}$ be a one-hot vector for the $j$th step of Tree-LSTM composition, where the hat denotes imperfect target labels induced by the PRPN's prediction. The parsing loss is defined as:
\begin{equation}
J_\text{parse}= - \sum\nolimits_j\sum\nolimits_i\widehat t_i^{(j)}\log p_i^{(j)}
\end{equation}
where $\bm p^{(j)}$ is the probability predicted by the Tree-LSTM model. The subscript $i$ indexes the $i$th position among in $1,\cdots,N_j-1$, where $N_j$ is the number of nodes in the $j$th composition step.

The overall training objective $J$ is a weighted combination of the loss of the downstream task and the parsing loss, i.e.,  $J=J_\text{task} + \lambda J_\text{parse}$. After step-by-step training, we perform \textit{policy refinement} by optimizing $J_\text{task}$ with ST-Gumbel, so that the Tree-LSTM can improve its policy based on a semantically oriented task.

It should be emphasized that how the Tree-LSTM model builds the tree structure differs between step-by-step training and ST-Gumbel training. For SbS training, we assume an imperfect parsing tree is in place; hence the Tree-LSTM model exploits existing partial structures to predict the next composition position. For ST-Gumbel, the tree structure is sampled from its predicted probability, enabling our model to explore the space of trees beyond the given imperfect tree.

\begin{table*}
	\label{LLM:tab:parsing}
	\small
	\centering
	\begin{tabular}{l | c c c| c c c} 
		\hline
		\multicolumn{1}{l|}{} & \multicolumn{3}{ c |}{ w/o Punctuation } & \multicolumn{3}{c}{w/ Punctuation} \\
		\cline{2-7}
		Model                       & Mean $F$      & Self-Agr & RB-Agr  &  Mean $F$     & Self-Agr  &  RB-Agr \\
    	\hline
		Left-Branching              & 20.7              & -             & -             & 18.9          & -             & -    \\
		Right-Branching             & \textbf{58.5}     & -             & -             & 18.5          & -             & -   \\
		Balanced-Tree               & 39.5              & -             & -             & 22.0          & -             & -   \\
		\hline
		ST-Gumbel                   & 36.4              & 57.0          & 33.8          & 21.9          & 56.8          & \textbf{38.1} \\
		PRPN                        & 46.0              & 48.9          & 51.2          & 51.6          & 65.0          & 27.4   \\
		Imitation (SbS only)        & 45.9              & 49.5          & 62.2          & 52.0          & \textbf{70.8} & 20.6   \\
		Imitation (SbS+refine)    & \,\,53.3$^\dag$   & \textbf{58.2} & \textbf{64.9} & \,\,\textbf{53.7}$^\dag$ & 67.4          & 21.1   \\
		\hline
	\end{tabular}	\caption{Parsing performance with and without punctuation. Mean $F$ indicates mean parsing $F$-score against the Stanford Parser (early stopping by $F$-score). \textbf{Self-Agr, RB-Agr:} Self-agreement and agreement with the right-branching baseline across multiple runs. $\dag$ indicates a statistical difference from the corresponding PRPN baseline with $p < 0.01$, paired one-tailed bootstrap test.}

	\label{LLM:tab:parse_res}
\end{table*}

\subsection{Experimental Results}
We train our model on the AllNLI dataset and evaluate it on the MultiNLI development set, following experimental settings in \cite{replication}.

Table \ref{LLM:tab:parse_res} shows the parsing $F$-scores against the Stanford Parser. The ST-Gumbel Tree-LSTM model and the PRPN were run five times with different initializations, each known as a trajectory.
For imitation learning, given a PRPN trajectory, we perform SbS training once and then policy refinement for five runs.
Left-/right-branching and balanced trees are also included as baselines. 

\paragraph{Parsing results with punctuation.} 
It is a common setting to keep all punctuation for evaluation on the AllNLI dataset \cite{replication}. 
In such a setting, we find that the Tree-LSTM, trained by ST-Gumbel from random initialization, does not outperform balanced trees, whereas the PRPN outperforms it by around 30 points. Our PRPN replication results are consistent with \cite{replication}.
Our first stage in imitation learning (SbS training) is able to successfully transfer the PRPN's knowledge to the Tree-LSTM, achieving an $F$-score of 52.0, which is clearly higher than the 21.9 achieved by the Tree-LSTM trained with ST-Gumbel alone, and even slightly higher than the PRPN itself. 
The second stage, policy refinement, achieves a further improvement in unsupervised parsing, outperforming the PRPN by 2.1 points. 

We also evaluate the self-agreement by computing the mean $F$-score across 25 runs for policy refinement and five runs for other models.
We find that our imitation learning achieves improved self-agreement in addition to improved parsing performance.

\paragraph{Parsing results without punctuation.}
We are interested in investigating whether punctuation makes a difference on unsupervised parsing.
In the setting without punctuation, our imitation learning approach with policy refinement outperforms the PRPN by a larger margin (7.3 $F$-score points) than in the setting with punctuation.
But surprisingly, strictly right-branching trees are a very strong baseline in this setting, achieving the best parsing performance overall. The PRPN cannot outperform the right-branching baseline, even though it uses a right-branching bias in its tree inference procedure.
By way of explanation, we assume that the syntactic trees we compare against (given by the Stanford parser) become more right-branching if punctuation is removed. A simple example is the period at the end of the sentence: this is always attached to a high-level constituent in the correct tree (often to Root), while right-branching attaches it to the most deeply embedded constituent. So this period is always incorrectly predicted by the right-branching baseline, if punctuation is left in.

To further elucidate this issue, we also compute the agreement of various models with a right-branching baseline. 
In the setting without punctuation, the PRPN sets an initial policy that agrees fairly well with right-branching, and this right-branching bias is reinforced by imitation learning and policy refinement.
However, in the setting with punctuation, the agreement with right-branching changes in the opposite way.
We conjecture that right-branching is a reason why our imitation learning achieves a larger improvement without punctuation. 
Right-branching provides a relatively flat local optimum so that imitation learning can do further exploring with a low risk of moving out of it.

\begin{figure*}
\centering
	\includegraphics[width=1.0\linewidth]{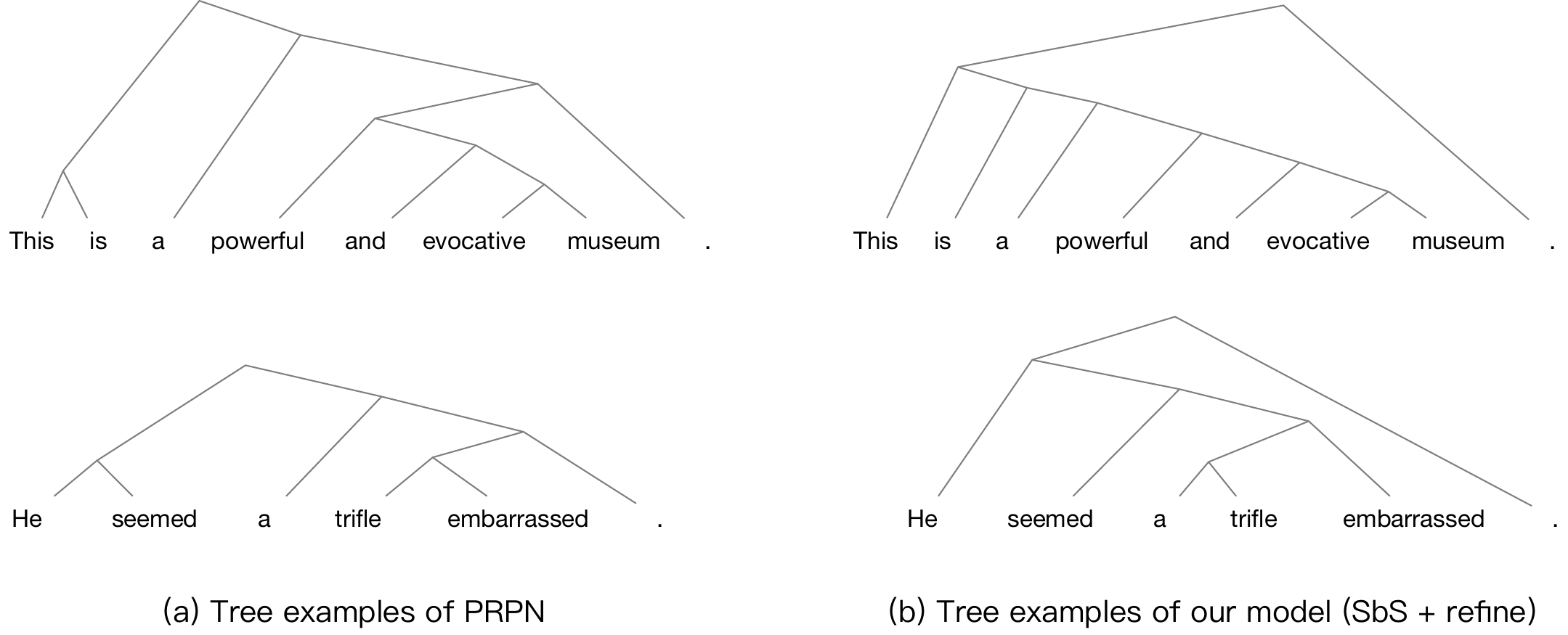}
	\caption{Parse tree examples produced by the PRPN and our model (SbS + refine).}
	\label{LLM:fig:tree_samples}
\end{figure*}

\paragraph{Case study.} In Figure~\ref{LLM:fig:tree_samples}, we present a few examples of parse trees generated by the PRPN and by our model (SbS + refine). 

As can be seen, our model is able to handle the period correctly in these examples. Although this could be specified by hand-written rules \cite{diora}, it is in fact learned by our approach in an unsupervised manner, since punctuation marks are treated as tokens just like other words, and our training signal gives no clue regarding how punctuation marks should be processed.
Moreover, our model is able to parse the verb phrases more accurately than the PRPN, including \textit{is a powerful and evocative museum} and \textit{seemed a trifle embarrassed}. 

\subsection{Summary}
In this section, we presented a neuro-symbolic approach to syntactic tree reasoning. A symbolic parser explicitly models tree-building operations, based on which a neural classifier predicts the natural language inference (NLI) label. The symbolic parser is pretrained with the PRPN model step by step, and then further learns to refine its tree-building process by Gumbel-softmax. Results show that learning from NLI improves the unsupervised parsing performance. 

Another important syntactic structure of language is the chunks, roughly speaking, phrases. Detecting chunks is fundamental to natural language understanding and can be used in downstream  tasks, such as logical reasoning~\cite{wu2021weakly}. In our preliminary work~\cite{chunk}, we have designed a hierarchical recurrent neural network (HRNN) that models word--phrase composition and phrase--sentence composition by a switching gate. We have also successfully pretrained such an HRNN by unsupervised parsing information, and we plan to apply our neuro-symbolic framework to learn chunking information in a downstream NLP task (such as summarization).

\section{Information Extraction Reasoning}\label{LLM:sec:IE}
Information extraction (IE) aims to extract structured ``machine
readable'' information from the unstructured or semi-structured  documents presented in free texts. IE is a fundamental task in NLP and involves multiple tasks such as text classification~\cite{pang2002thumbs,zhou2016attention-based,Wang2012Baselines,Zhang15-cnnText}, template filling~\cite{du2021template}, semantic parsing~\cite{buys}, and relation extraction~\cite{zelenko2003kernel}.

To highlight the key points, the natural language is usually informative and the commonsense that is already known for humans is oftentimes omitted. The neural networks are a black box and are thus difficult to combine the human knowledge in a natural way. Therefore, neural networks need to perform some reasoning on the texts during the information extraction. In this section, we will introduce the advances of neuro-symbolic reasoning methods on two fundamental IE tasks, i.e., text classification and document parsing.

\subsection{Locating Evidence for Text Classification}
Text classification aims to categorize a piece of text into predefined classes.\footnote{Part of this chapter was published in~\cite{jumper}, available at \url{https://www.ijcai.org/proceedings/2018/589}, reused with permission. \copyright2018  International Joint Conference on Artificial Intelligence. Changes are made to fit this book chapter.} Previous work mainly focuses on the ultimate performance of a task (e.g., classification accuracy). For example, Kim~\cite{Kim14sent} builds several variants of convolutional neural networks for sentiment classification. However, we are more preferred to the text classifier that can not only predict the target category but also indicate the evidence for that prediction, which is in fact important in real industrial applications for debuggability, believability, and interpretability~\cite{criticism}. 

To achieve the above goal, we propose a neuro-symbolic framework, coined Jumper, that models text classification as a sequential decision process, inspired by the cognitive process of humans. When people read text, we look for clues, perform reasoning, and obtain information from text. Jumper mimics this process by reading the text in a sentence-by-sentence manner with a neural network. At each sentence, the model makes decisions (also known as \textit{actions}) based on the input, and at the end of this process, it would have some ``understanding'' of the text. This work also follows the general neuro-symbolic framework in Figure~\ref{fig:overview}, since decision actions are intermediate thinking steps towards the ultimate classification.  

To be more specific, Jumper focuses on paragraph-level text classification, which may have more semantic changes than the sentence level. When our neural network reads a paragraph, it is assumed to have a default value ``\texttt{None}'' at the beginning. At each decision step,  a sentence of the paragraph is fed to the neural network; the network then decides if it is confident enough to ``jump'' to a non-default value. We impose a constraint that each jump is a finalized decision, which cannot be updated in the future. 

An intriguing consequence of the one-jump constraint is that it forces our model to be serious about both when to predict and what to predict. This is because a paragraph does not contain a special symbol indicating the end of the paragraph. If our model defers its decision later than it could have made an accurate enough prediction, it takes a risk of not being able to predict. On the other hand, if the model predicts too early, it takes a risk of low accuracy. By optimizing the expected reward in reinforcement learning, the model learns how it can make decisions at an ``optimal'' time, and thus providing the additional classification evidence.

\begin{figure}[!t]\centering
\includegraphics[width=0.5\linewidth]{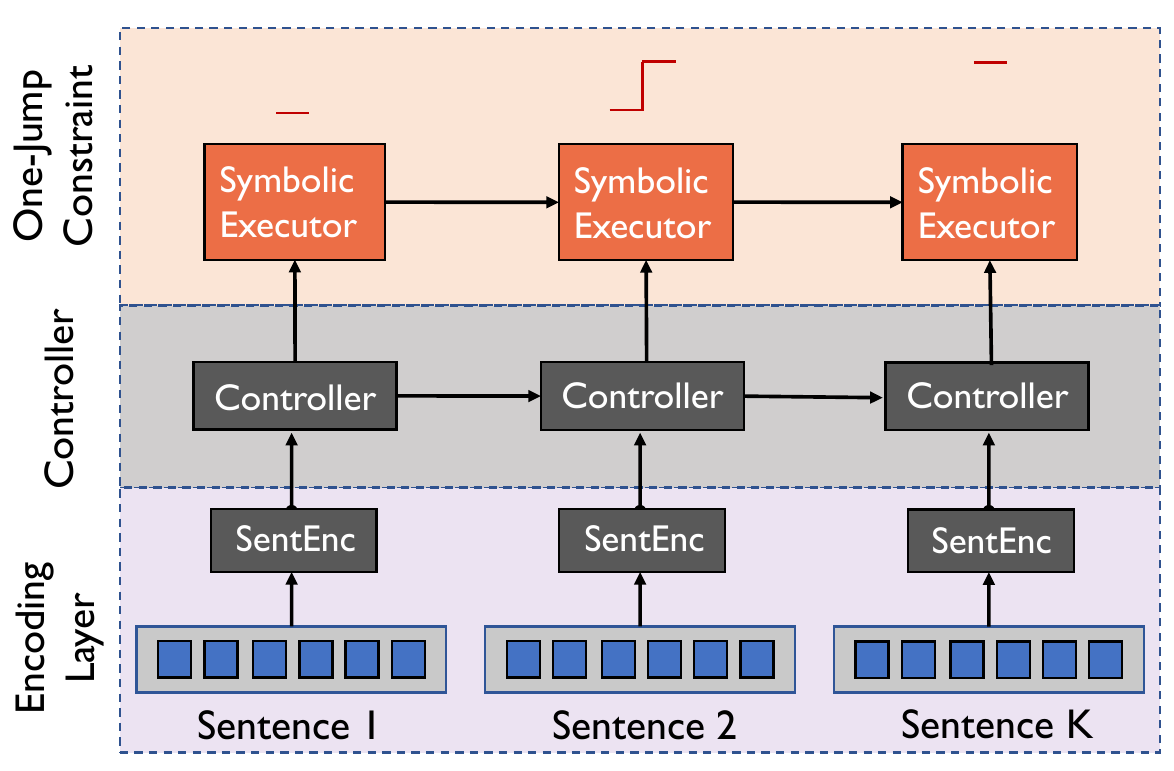}
\caption{The Jumper framework.}\label{fig:jumper}
\end{figure}

\subsubsection{The Jumper Model} 
Figure~\ref{fig:jumper} shows the overall framework of our approach. We first segment the paragraph into sub-sentences; each could be thought of as a basic unit for some “proposition,” and is fed to our model in order. There are three main components in our neural network, including a sentence encoder, a controller, and a symbolic executor. The sentence encoder encodes the semantic features of words
in a sentence into a fixed-dimensional vector space. The controller, essentially a recurrent neural network
(RNN), is built upon sentence encoders, and takes actions
(“jumps”) when appropriate. And the symbolic executor maintains the decisions that have been made, and ensures consistency according to hard constraints that we impose. The classification results are the final output of the symbolic executor. The rest of this subsection will elaborate these three components in detail.

\paragraph{Sentence encoder (SentEnc).} We use a convolutional neural network (CNN) as the sentence encoder \cite{Kim14sent}. 
CNN applies a set of sliding windows to the concatenation of neighboring words to extract local features, which are aggregated by max pooling to represent semantic features for individual sentences. 

\paragraph{Controller.} Based on encoded sentence features, the controller of Jumper takes corresponding actions, as in a sequential decision process. Inside the controller are two submodules: (1) an RNN fuses the current input and previous sentences, maintaining dependency
 over the entire history; and (2) a policy network (PolicyNet) makes a decision for the current step.

\paragraph{Symbolic executor.}
The symbolic executor of JUMPER maintains the decisions that have been made until the current sentence. It performs symbolic calculations for the final decision (i.e., the classification category). In symbolic executor, we propose a one-jump constraint that allows at most one non-default prediction (not ``\texttt{None}'') for each slot. Let $\bm a_t\in\mathbb{R}^{N+1}$ denote PolicyNet's decision  after processing the $t$th sentence, and $\bm s_t^{(i)}\in\{0,1\}^{N_i+1}$ be the one-hot representation of the symbolic layer's state for slot $i$ at the $t$th sentence. Formally, the one-jump constraint is given by
\begin{align}
\bm s_t^{(i)} &=\bm s_{t-1}^{(i)}\cdot \boldsymbol{1}_{\{s_{t-1}^{(i)}\ne\texttt{None}\}} +\bm a_t\cdot \boldsymbol{1}_{\{s_{t-1}^{(i)}=\texttt{None}\}}
\end{align}
where $ \boldsymbol{1}_{\{\cdot\}}  $ is an indicator function that yields 1 when its argument is true, and 0 otherwise. In other words, the state has to be the same as its previous one if it is not ``\texttt{None}''; but on the other hand, JUMPER can remain in ``\texttt{None}'' for the entire paragraph if the information does not exist. The final result, which we use to compare with groundtruth, is the symbolic layer's state after processing the last ($T$th) sentence, i.e.,  $s_T^{(i)}$.

\paragraph{Learning.} As most of the neuro-symbolic frameworks encounter, the main learning difficulty of Jumper is the lack of step-by-step supervision. That is, we assume the labels contain only ultimate results for each slot, but no information for the appropriate position to jump. Admittedly, life will be easier if we have fine-grained annotations regarding which step to predict for each slot, but they are costly and labor-intensive to obtain. We therefore apply reinforcement learning to train our Jumper framework. We define a reward by comparing the model's prediction and the groundtruth. The training objective is to maximize the expected reward over sampled actions. Therefore, Jumper is optimized towards the maximum of the obtained rewards.

\begin{figure}[!t]\centering
\includegraphics[width=0.75\linewidth]{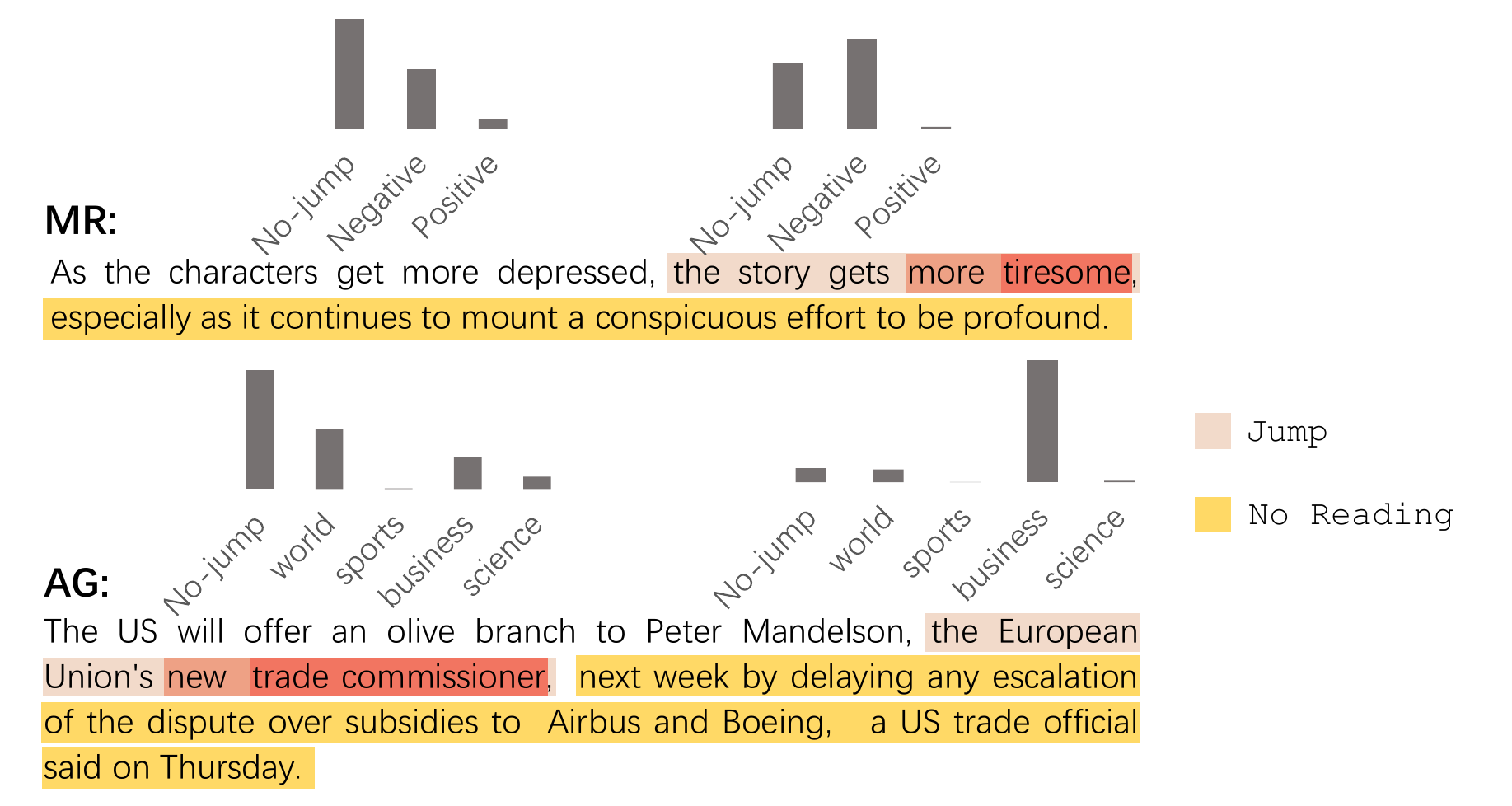}
\caption{The case study that illustrates the decision process of Jumper. We also show the histogram of decision distributions and the heatmaps of word importance in MR and AG samples.}\label{fig:showcase}
\end{figure}

\subsubsection{Results} 

The Jumper framework is evaluated in multiple ways including classification accuracy and the accuracy of jumping at the evidence sentence. The experiments show that Jumper  achieves comparable or better performances over the previous methods on the movie review dataset, AG news corpus, and also self-collected occupational injury dataset. Further, we also observe that Jumper could find the optimal decision step in the simulation experiments~\cite{jumper-neurocomputing}.

Besides the quantitative analysis, we also qualitatively show several examples of the decisions made by the neural network in Figure~\ref{fig:showcase}. The samples in the movie review (MR) and AG news corpus (AG) datasets reflect the general characteristics of natural language, i.e., the information of interest is located over a wider range. The network makes a prediction as long as it sees enough evidence (e.g., ``trade commissioner'' for the business domain). By backtracking the word-level rationales (a gradient-based attribution technique like~\cite{sundararajan2017axiomatic}), we find words like  ``trade commissioner'' and ``tiresome''  play a more important role in the decision making. In these cases, the model does not need to read future sentences, which is more efficient than reading the entire paragraph.

\subsection{Object-Oriented Programming for Document Parsing}
Mapping a document into a structured ``machine readable" form is a canonical and probably the most effective way for document parsing~\cite{berant2014semantic}.\footnote{Part of this chapter was published in~\cite{OONP}, available at \url{https://aclanthology.org/P18-1253/}, licensed under \href{https://creativecommons.org/licenses/by/4.0/}{CC BY 4.0}. \copyright2018 Association for Computational Linguistics. Changes are made to fit this book chapter.}  There are several recent efforts on designing neural network-based learning machines for this purpose, which can be roughly categorized into two groups:  1) sequence-to-sequence models with the neural net as the black box~\cite{SP}, and 2)  neural networks as a component in a pre-designed statistical model~\cite{Zeng14-relationClassification}. Both categories are hindered in tackling document with complicated structures,  by either the lack of effective representation of knowledge or the flexibility in fusing them in the model. 

\begin{figure}[!t]
	\centering
	\includegraphics[width=1.0\linewidth]{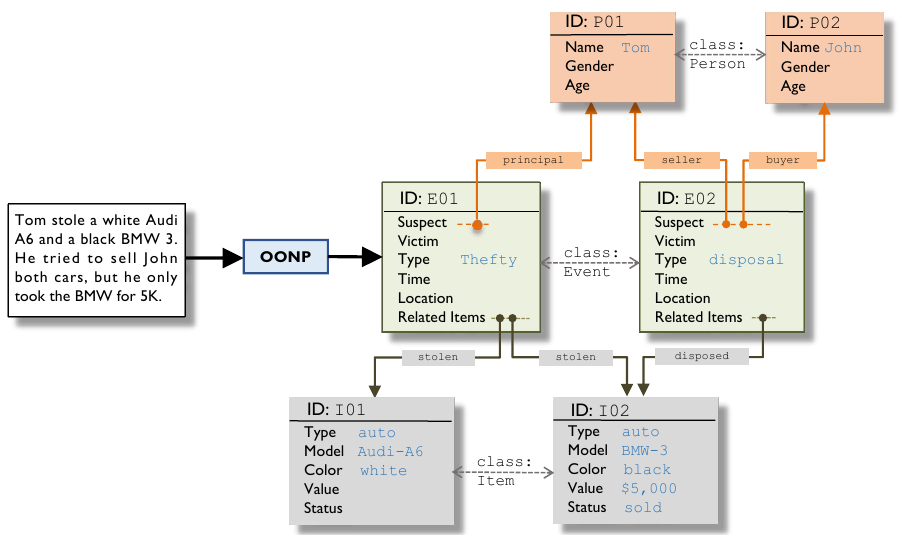}
	\caption{Illustration of OONP on a parsing task.}\label{fig:oonp}
\end{figure}

Towards solving this problem,  we proposed \textit{ Object-Oriented Neural Programming} (OONP),  a framework for semantically parsing in-domain documents. As shown in Figure~
\ref{fig:oonp}, OONP maintains an object-oriented data structure, where objects from different classes are to represent entities (people, events, items, etc.) which are connected through links with varying types.  Each object encapsulates internal properties (both symbolic and differentiable),  allowing both neural and symbolic reasoning over complex structures and hence making it possible to represent rich semantics of documents. OONP also follows our general framework (Figure~\ref{fig:overview}), as the object-oriented structures (both symbolic and distributed ones) are generated, updated, and employed during the document parsing process. To summarize, an OONP parser is based on neural networks,  but it has sophisticated architectures and mechanisms designed for taking and yielding discrete structures, hence nicely combining symbolism (for interpretability and formal reasoning) and connectionism (for flexibility and learnability).

\subsubsection{The OONP Model}
\begin{figure}[!t]\centering
\includegraphics[width=0.8\linewidth]{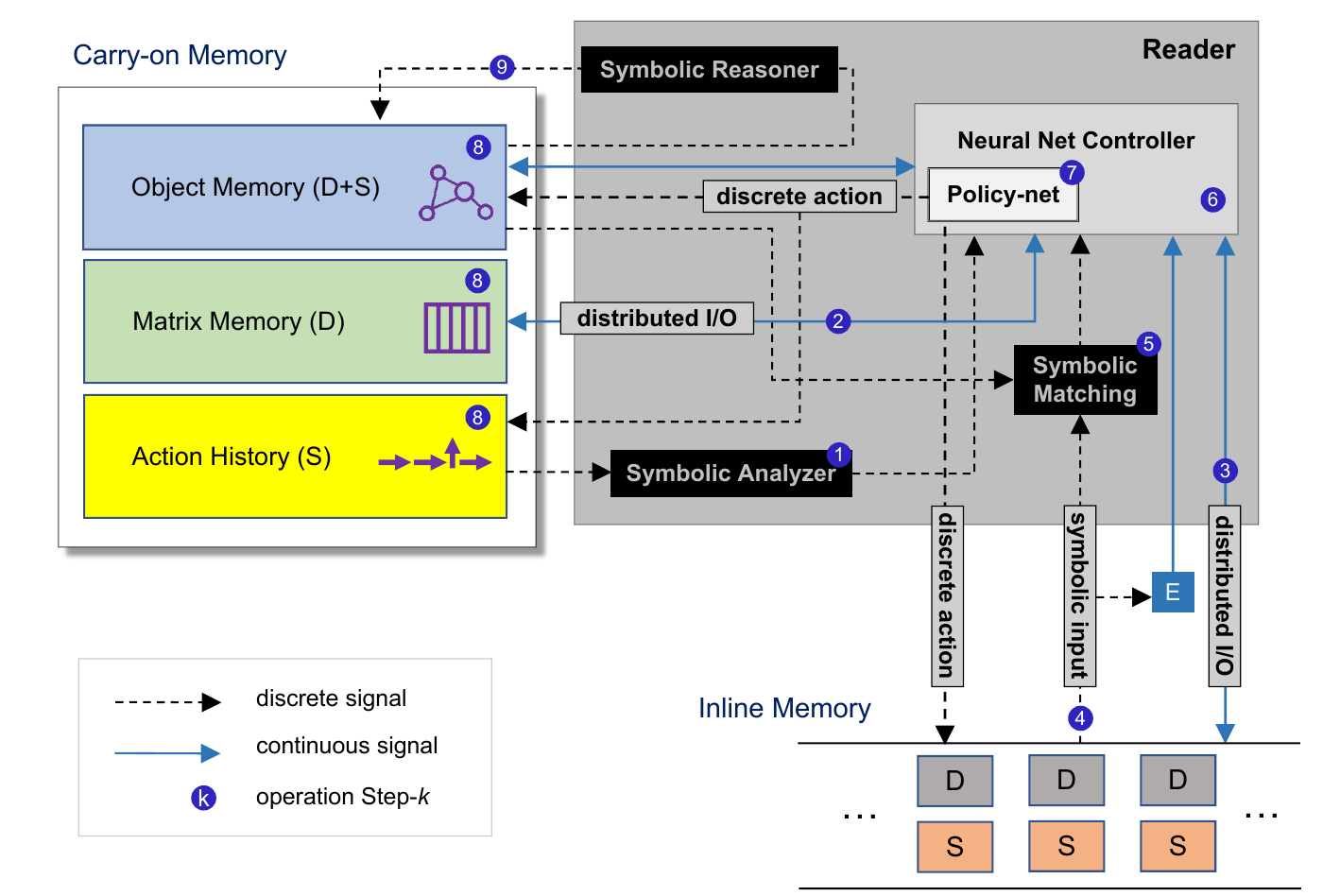}
\caption{The OONP model.}\label{fig:oonp-model}
\end{figure}

An OONP parser (as illustrated through the diagram in Figure \ref{fig:oonp-model}) consists of a reader equipped with read/write heads,  an inline memory that represents the document,  and a carry-on memory that summarizes the current understanding of the document at each time step.  For each document to parse, OONP first preprocesses it and puts it into the inline memory, and then the reader controls the read heads to sequentially go through the inline memory (for possibly multiple times) and at the same time to update the carry-on memory. The major components of OONP are described in the following: 

\paragraph{Object-oreinted memory.}  We have two types of memory,   carry-on memory and inline memory. The carry-on memory is designed to save the state in the decision process and summarize current understanding of the document based on the text that has been read.  The carry-on memory has three compartments: 
\begin{itemize}
\item Object memory: Object memory stores an object-oriented representation of document. Each object is an instance of a particular class, which specifies the internal structure of the object, including internal properties, operations, and how this object can be connected with others (i.e., links or relations).
\item Matrix memory: A matrix-type memory with a fixed size, for differentiable read/write by the controlling neural net~\cite{graves2014neural}. In the simplest case, it could be just a vector as the hidden state of a conventional Recurrent Neural Network (RNN);
\item Action history: It maintains the entire history of actions made during the parsing process.
\end{itemize}

Besides, the inline memory stores the relatively raw representation of the document that follows the temporal structure of the text. Basically, the inline memory is an array of memory cells, each corresponding to a pre-defined language unit (e.g., word) in the same order as they are in the original text. Each cell can have a distributed part and a symbolic part, designed to store 1) the preprocessing result of text from different models, and 2) certain output from the reader, for example, from previous reading rounds.

\paragraph{Reader.} The reader is the control center of OONP, coordinating and managing all the operations of OONP. More specifically, it takes the input of different forms (reading), processes it (thinking), and updates the memory (writing). The reader contains a neural net controller (NNC) and multiple symbolic processors, and NNC also has a policy net as a sub-component. Similar to the controller in the neural Turing machine~\cite{graves2014neural}, NNC is equipped with multiple read heads and write heads for differentiable read/write over matrix memory and (the distributed part of) inline memory, with possibly a variety of addressing strategies~\cite{graves2014neural}. The policy net, however, issues discrete outputs (i.e., actions), which gradually builds and updates the object memory in time. The symbolic processors are designed to handle information in symbolic form from the object memory, the inline memory, the action history, and the policy net. 

\paragraph{The workflow of OONP.}
The reader is the control center of OONP, which manages all the (continuous and discrete) operations in the OONP parsing process.  All the components in the reader are coupled through intensive exchange of information as shown in Figure~\ref{fig:oonp-model}. Below is the workflow of the information processing at
time $t$.

{ \small
\begin{itemize}
\item \textbf{STEP-1:} Let the symbolic analyzer (a processor in the reader) to check the action history to construct some symbolic features for the trajectory of actions;
\item \textbf{STEP-2:} Access the matrix memory to get vectorial representations for time $t$,  denoted as  $s_t$; 
\item \textbf{STEP-3:} Access inline memory to get the symbolic representation $x_t^{(s)}$ (through location-based addressing) and distributed representation $x^{(d)}_t$ (through location-based addressing and/or content-based addressing); 
\item \textbf{STEP-4:}  Feed $x_t^{(d)}$ and the embedding of $x_t^{(s)}$  to the neural net controller to fuse with $s_t$;
\item \textbf{STEP-5:}  Get the  candidate objects and let them meet $x_t^{(d)}$ through the processor  ``Symbolic Matching''  for the matching of them on symbolic aspect;
\item \textbf{STEP-6:}  Get the  candidate objects (some may have been eliminated by $x_t^{(s)}$) and let them meet the result of STEP-4  in the neural net controller;
\item \textbf{STEP-7:} The policy net combines the result of STEP-6 and STEP-5,  to issue actions; 
\item \textbf{STEP-8:} Update all the memories with actions on both symbolic and distributed representations; and
\item \textbf{STEP-9:} Put object memory through the processor ``Symbolic Reasoner'' for some high-level reasoning and logic consistency. 
\end{itemize}}

\subsubsection{Results}
The OONP model is designed to generate an ontology for a given document through a decision process with symbolic actions. We implement OONP for parsing  Chinese police reports (brief description of criminal cases written by police officers).   We consider a corpus of 5,500 cases with a variety of crime categories,  including theft, robbery, and drug dealing.  The ontology we designed for this task mainly consists of a number of person objects and item objects connected through an event object with several types of relations. Each type of the objects has several specific properties. For example, a person object has three internal properties: name (string), gender (categorical), and age (number), and two types of external links (suspect and victim) to an event object. We use 4,250 cases for training, 750 for validation, and another 750 for test. We consider the following four metrics in comparing the performance of different models, including assignment accuracy, category accuracy,  ontology accuracy, and ontology accuracy-95. In particular, the metric of ontology accuracy-95 measures the proportion of the predicted ontology that gets 95\% consistency with the ground truth. Overall, these metrics measure the accuracy of the model in making discrete decisions as well as generating the final ontology. 

As shown in Table~\ref{tab:oonpresultsx}, the Bi-LSTM baseline struggles to achieve around 73\% assignment accuracy on test set. Another baseline ENTITYNLM~\cite{ji-etal-2017-dynamic} largely boosts the assignment accuracy by introducing the concept of the entity in the neural network. Taking one step further, OONP leverages the object memory to facilitate the parsing of a document, which results in better document parsing performance, especially in the ontology accuracy. When we put symbolic representations and operations into the reader, as shown in the result of OONP (structure), the performance is again significantly improved on all four metrics, further demonstrating the superiority of OONP in parsing complex documents.

\begin{table}[!t]
	\centering
	\resizebox{0.9\linewidth}{!}{
		\begin{tabular}{lcccc}
			\toprule
			Model &   Assign Acc. (\%)  & Type Acc. (\%)  &   Ont. Acc. (\%)  & Ont. Acc-95 (\%) \\
			\midrule
			Bi-LSTM (baseline) &  73.2 $\pm$ 0.58 &- & 36.4$\pm$ 1.56 & 59.8 $\pm$ 0.83\\
			ENTITYNLM (baseline) & 87.6 $\pm$ 0.50 & 84.3 $\pm$ 0.80  & 59.6 $\pm$ 0.85  & 72.3  $\pm$ 1.37 \\
			OONP (neural)  & 88.5 $\pm$ 0.44 & 84.3 $\pm$ 0.58  & 61.4 $\pm$ 1.26 & 75.2 $\pm$ 1.35 \\
			OONP (structured) &  91.2 $\pm$ 0.62 & 87.0 $\pm$ 0.40 & 65.4 $\pm$ 1.42 & 79.9 $\pm$ 1.28 \\
			\bottomrule
		\end{tabular} }	\caption{Model performance on parsing police reports.} 

		\label{tab:oonpresultsx}
	\end{table}

\subsection{Summary}
 In this section, we introduce two neuro-symbolic methods for information extraction reasoning. To extract the information of interest (e.g., the category and the rationale) from a sentence, we introduce the Jumper model that leverages the one-jump constraint to enforce the agent to yield symbolic classification evidence during the decision process. To extract information from a document, we propose the OONP model that supports a variety of forms (both symbolic and differentiable) for representing the state and the document with
 a pre-designed object-oriented data structure.

\section{Rule Reasoning}\label{LLM:sec:rule}
One of the distinctive features of human intelligence is the ability to summarize abstract rules from historical experience, which we call the ability of rule learning~\cite{lake2017building}.\footnote{Part of this chapter was published in~\cite{ruler}, available at \url{https://www.ijcai.org/proceedings/2022/593}, reused with permission.  \copyright2022 International Joint Conference on Artificial Intelligence. Changes are made to fit this book chapter.} Cognitive scientists found that seven-month-old infants can summarize the simple rules contained in the natural language~\cite{marcus1999rule}. However, the prevailing artificial intelligence approaches pale in comparison to the human ability in rule learning. As the most effective artificial intelligence method, deep neural networks tend to learn statistical characteristics of training samples rather than discrete, symbolic rules. Therefore, deep neural networks have major defects in interpretability, generalization, and data utilization efficiency, which greatly hinders real-world applications.

The current research on rule learning and reasoning of deep neural networks is mainly limited to structured rules (i.e., the rules that are organized with the fixed format). For example, researchers regard the transformation relationship of relation triples in the knowledge graph as rules, and then carry out rule learning and knowledge reasoning~\cite{yang2017differentiable}. However, the above studies rely on manual extraction of structural rules, which requires heavy human labor. At the same time, real-world data and the underlying rules hardly have a fixed structure, research based on structured rules cannot meet the requirements of the real world for rule learning capabilities. Therefore, how to construct a neural network method with the ability of rule learning and reasoning for unstructured rules is an important but challenging research problem.

\begin{figure*}[!t]
	\centering
	\includegraphics[width=0.6\linewidth]{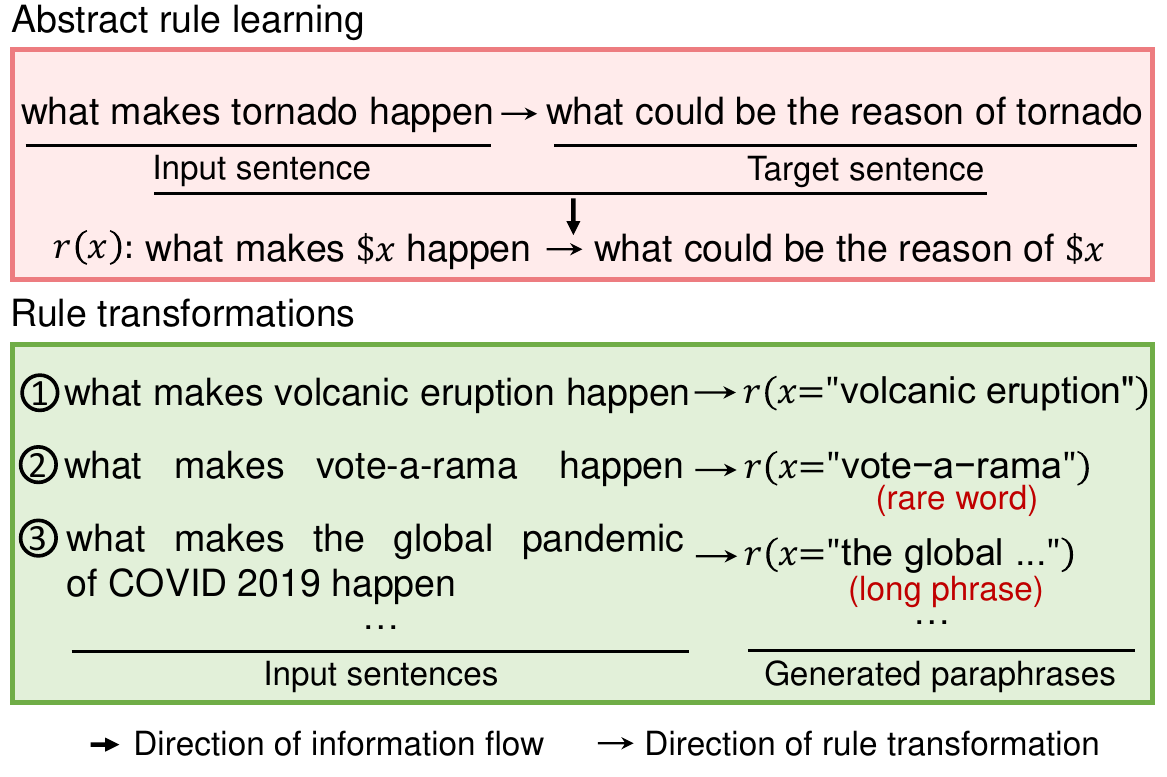}
	\caption{An example of the learning and transformation of an abstract rule $r$. $\$z$ is a symbol (placeholder) that can be assigned to various words during rule transformations.
	}
	\label{fig:ruleridea}
\end{figure*}

Considering that the paraphrase corpus contains a wealth of unstructured paraphrase rules, the study on paraphrase rules could be a main frontier of the rule learning field. The paraphrase generation task refers to generating sentences that are semantically consistent with the input sentences but expressly different~\cite{Barzilay-rules,upsa}. For example, given the sentence ``why does a volcano erupt'', a possible paraphrase is ``what is the reason of the eruption of the volcano''. This example can be characterized by the rule ``what does $z$ happen $\rightarrow$ what is the reason of $z$'', where $z$ represents a placeholder in the rule (see Figure~\ref{fig:ruleridea} for more examples). Compared with the relational transformation rules in the knowledge graph, the paraphrase rules are unstructured and have rich semantics and diversity, making it difficult for traditional matching-based methods to utilize them efficiently~\cite{quirk2004monolingual,wang2019a,prakash2016neural,gupta2018deep,separator}. Except for a few artificially induced paraphrase rules, a large number of paraphrase rules are still buried in the paraphrase corpus~\cite{zhao2009application}. Thus, the paraphrase corpus provides a unique opportunity to test the rule learning capacity of neural networks.
 
In this work, we integrate the advantages of the rule-based and neural-based methods into RULER, an interpretable framework that generates paraphrases by abstract \textit{RUle LEaRning}. The abstract rule is the generalizable transformation from perceptual inputs to desired outputs. Our key idea is to learn abstract rules underlying paraphrasing that benefit the paraphrase generation process. From the view of our neuro-symboli framework~(Figure~\ref{fig:overview}), these rules can be thought of as the intermediate thinking process for generating a paraphrase.
Specifically, we first propose an evaluation metric to indicate the rule generalizability. Next, a rule learner learns to generate rules by maximizing the expected rewards (i.e., rule generalizability).
Then, we select a suitable rule from the learned rules for the coming sentence and apply the symbolic rule-based transformation. Finally, a neural network takes the transformed sentence as a reference to generate the paraphrase.

\subsection{The RULER Framework}
In this section, we will first introduce the definition of the abstract rule and then elaborate the proposed rule learning process. Finally, we describe how to leverage the learned rules to generate paraphrases (i.e., the rule-enhanced generation process).

\paragraph{Abstract rule.} The abstract rule was first introduced to study the language acquisition of infants~\cite{marcus1999rule}. It is defined as a simplified grammar that is highly abstract. For example, the rule ``ABB'' stands for three-word sentences where the last two words are the same and differ from the first one. The token ``A'' and ``B'' are variables (placeholders) that can be assigned to any word. In this work, we use an abstract rule to represent a type of paraphrase transformation. Let $\mathbb{W}$ be the word vocabulary and  $\mathbb{Z}$ be the vocabulary of the symbols (placeholders) that can be assigned to various words. An abstract rule $r=\bm w\rightarrow \bm v$ is an implication corresponding to the conditional statement: ``if $\bm w$ is matched, then produce $ {\bm v}$'', where $\bm w=[w_1,\dots,w_{L_w}]$ and  ${\bm v}=[v_1,\dots,v_{L_v}]$ are the abstract patterns of the input sentence and the corresponding paraphrase, respectively. $L_w$ and $L_v$ are the lengths of tokens of $\bm w$ and $\bm v$. Note that $\bm w$ and $\bm v$ are not sentences because they commonly contain symbols. An example rule that contains two symbols is ``what are the differences between $x_1$ and $x_2\rightarrow$ how do $x_2$ and $x_1$ differ.'' The symbols stands for all the possible words that make the input and target sentences fluent. Therefore, the rule is highly abstract and thus generalizes well, even to  out-of-vocabulary words. 

Formally speaking, a rule $r$ describes the mapping from the input sentence to its paraphrase, which is symbolized by p $\rightarrow$ q, it is an if-then statement in which p is a hypothesis and q is a conclusion.
\begin{align}
\bm z &= [z_1, \dots,z_i, \dots, z_{L_z}], z_i \in \mathbb{Z},  i\in 1,2,\dots,L_z \\
r(x) &= \bm w\rightarrow \bm v =  [w_1,\dots,w_{L_w}] \rightarrow  [v_1,\dots,v_{L_v}]  \\
w_l,& v_k\in \mathbb{W}  \cup \bm z; l\in  1,2,\dots,L_w; k\in 1,2,\dots,L_v, 
\end{align}
where $\bm z$ stands for the set of symbols used in rule $r$. $L_z$ is the number of the symbols in rule $r$.

\begin{figure*}[!t]
	\centering
	\includegraphics[width=0.8\linewidth]{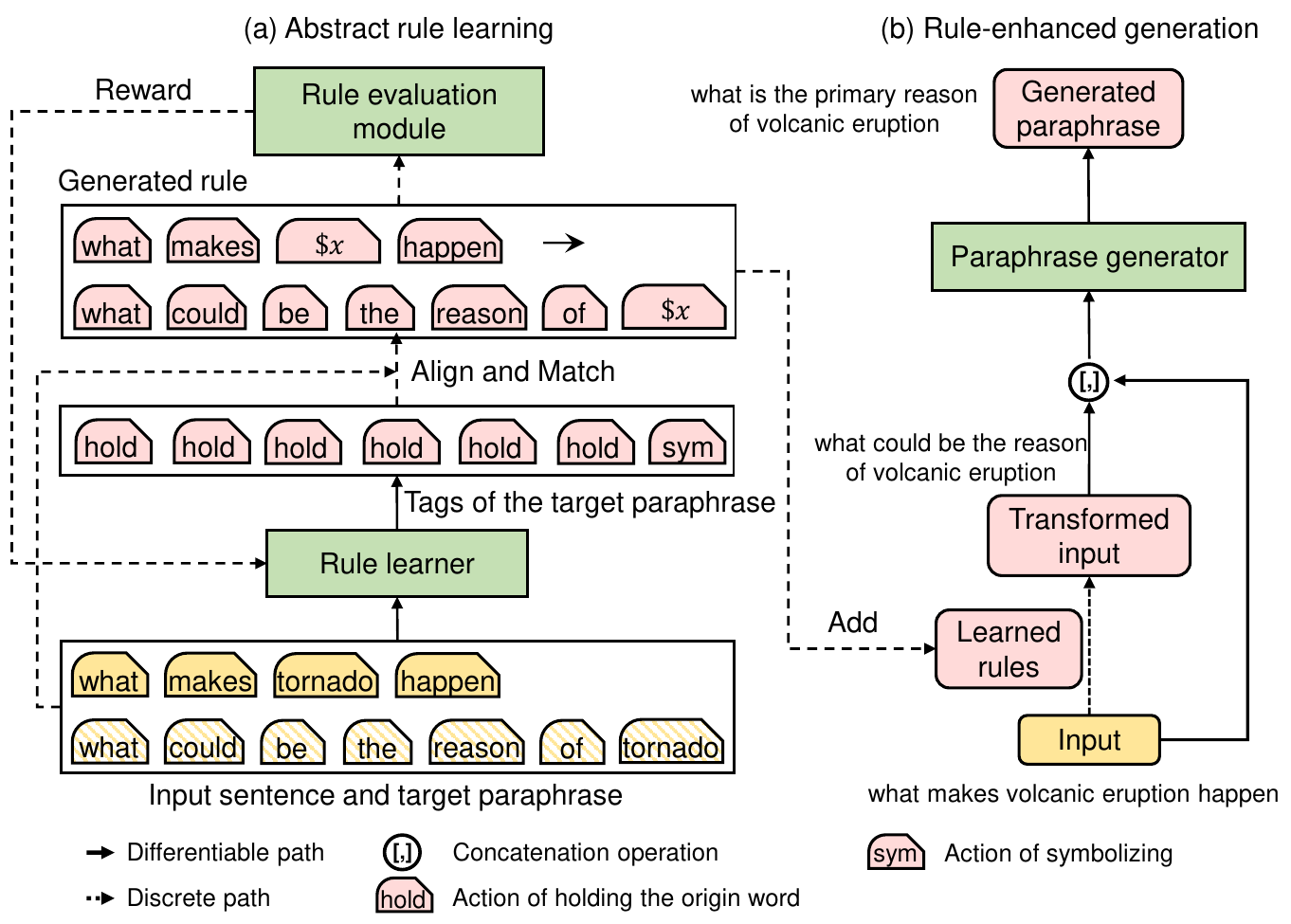}
	\caption{Overview of the RULER framework.
	}
	\label{fig:rulermodel}
\end{figure*}

\paragraph{Abstract rule learning.} As shown in Figure~\ref{fig:rulermodel}a, the rule learning process mainly involves a rule learner and a rule evaluation module. The rule learner takes a data point (including both the input sentence $\bm X$ and its paraphrase $\bm Y$) as input and generates a candidate rule underlying this data point. The rule evaluation module assesses the generalizability of the extracted rules, and thus guides the learning of the rule learner. We will first formulate the rule generation subtask and then elaborate on these two components in detail.

The rule generation subtask is to extract generalizable rules from a given data point. However, if we generate the rule expression token by token, the generative complexity is $O(| \mathbb{W}  \cup \mathbb{Z}|^{L_w+L_v})$, which will result in low search efficiency. Since the evaluation of each rule is time-consuming (explained later), it will take a long time to figure out a suitable rule for a data point, making it intractable to train a rule learner. In this work, we try to simplify the rule generation subtask by restricting the output space of the rule learner. 

Considering that the extracted rule $r$ is expected to characterize the general mapping of the given data point, we assume that the tokens of an abstract rule are either the words in the given data point or symbols.  Specifically, for each word $Y_{l}$ in the target paraphrase $\bm Y$, the rule learner predicts a tag $O_l$, deciding whether the word should be preserved (i.e., the action of holding) or symbolized (i.e., the action of symbolizing). 
\begin{align}
	\bm O = [O_1,\dots, O_{L_o}]= \text{RuleLearner}_{\bm \theta}(\bm X, \bm Y),
\end{align}
where $\theta$ denotes all model parameters and $O_l\in \{\tt{hold},\tt{symbolize}\}$. $\bm O$ is the output of the rule learner and $L_o$ is the length of the output $\bm O$. Note that, to have a one-to-one mapping between the output and the target paraphrase, the length of the output is the same as the target paraphrase. Based on the sequence of binary decisions of the rule learner, it is not difficult to align the symbols to the target paraphrase and the input sentence, as illustrated in Figure~\ref{fig:rulermodel}. 

\paragraph{Rule evaluation.}
Besides the huge search space, the lack of direct supervision is the second challenge of rule learning. In this work, we develop a rule evaluation module to automatically assess its quality. We consider the quality of a rule in two aspects, namely, the coverage of the rule and the usefulness of the rule for paraphrase generation. For the first aspect, the rule evaluation module tries to figure out how many data points that rule $r$ covers. The wider coverage of a rule, the better generalization it possesses. In particular, it calculates the coverage score by counting the samples from the training data that conform to the abstract rule. For the second aspect, the rule evaluation module tries to answer the question how many data points that rule $r$ can help to perform paraphrasing. We suppose that, if the sentence transformed by rule $r$ provides more information for paraphrase generation than the original sentence $\bm X$, rule $r$ is useful to $\bm X$. The rule evaluation module leverages a trained paraphrase generator to evaluate the difficulty of the paraphrasing based on the different inputs.

\paragraph{Rule-enhanced paraphrase generation.}
For a sentence with a suitable abstract rule, we presume that it is easier to generate its paraphrase from the transformed sentence than the original sentence. Thus, we aim to train a powerful generator (i.e., a Transformer model~\cite{transformer}) to perform paraphrasing based on the learned rules (Figure~\ref{fig:rulermodel}b), which we call the rule-enhanced generation process. In this process, we add the transformed sentence as the auxiliary input into the training data of the paraphrase generator in hopes of generating more diverse paraphrases.

\begin{table}
	\centering
	\resizebox{0.8\linewidth}{!}{
		\begin{tabular}{llcccc}
			\toprule
			Input                            &  \!\!\!\!\!How do i become an investment banker? What do they do  \\
			\multirow{2}*{Matched rule}      & \!\!\!\!\!How do i become $\$z$? What do they do $\rightarrow$ \\
			&  \!\!\!\!\!What is the best way to be  $\$z$    \\
			RULER &  \!\!\!\!\!What is the best way to be an investment banker \\
			SEPARATOR  &  \!\!\!\!\!How can i become an investment banker  \\
			\midrule		
			Input &   \!\!\!\!\!What is Taylor Swift 's nationality\\
			Matched rule     &  \!\!\!\!\!What is  $\$z$ 's nationality $\rightarrow$ What nationality is $\$z$  \\
			RULER &   \!\!\!\!\!What nationality is Taylor Swift   \\
			SEPARATOR  &  \!\!\!\!\!What is Taylor Swifts nationality and why  \\
			\midrule		
			\multirow{2}*{Input}     
			    & \!\!\!\!\!Do wild animals suffer immensely when hunted by a wild predator, or are\\
			    & \!\!\!\!\!there mechanisms at work (such as adrenaline) that alleviate any suffering \\
			Matched rule & {\tt \!\!\!\!\!None} \\
			\multirow{2}*{RULER}     
			    &  \!\!\!\!\!Does wild animal suffer immensely when hunted by a wild predator, or\\
				& \!\!\!\!\!are there any related mechanisms\\
			SEPARATOR  	 & \!\!\!\!\!Are wild animals more dangerous than wild animals	 \\
			\bottomrule
		\end{tabular}
	}
	\caption{Paraphrases generated by RULER and SEPARATOR.}

	\label{tab:rulerexample}
\end{table}

\subsection{Results}

We evaluate RULER on two widely used datasets, namely, the Quora question pairs and Wikianswers datasets. We adopt BLEU, iBLEU, and METEOR scores as automatic metrics to evaluate the generation performance. Among them, iBLEU is the BLEU score penalized by the similarity with the original sentence. It is more comprehensive for evaluation, and thus we take it as our major metric. We compare our method with multiple advanced models, e.g., Transformer~\cite{transformer}, LBoW~\cite{fulatentbow}, variational autoencoder (VAE)~\cite{gupta2018deep}, and SEPARATOR~\cite{separator}. Experiments indicate that RULER yields much better results than the other paraphrasing systems, including the previous state-of-the-art model (i.e., SEPARATOR). 

We also showcase several generated rules and corresponding paraphrases in Table~\ref{tab:rulerexample}. We see qualitatively that RULER produces more reasonable paraphrases than SEPARATOR in terms of both closeness in meaning and difference in expressions. In these examples, SEPARATOR just changes a few inconsequential words. By leveraging learned rules, RULER can make global modifications to the inputs. Meanwhile, RULER fails to provide suitable rules for some complicated sentences.

\subsection{Summary}
Abstract rule learning is the ability to find generalizable transformations from the perceptual input to the desired output~\cite{marcus1999rule}. The rule learning capacity of deep models has attracted recent interest in the field of artificial intelligence~\cite{arii,barrett2018measuring}. In this section, we explore it by a specific problem, i.e., paraphrase generation. Our proposed method, RULER, first learns the symbolic paraphrasing rules by reinforcement learning and then generates paraphrases by taking the rule-transformed input as an auxiliary input. Experiments show that the inclusion of symbolic rules in the paraphraser yields better generation results.

\section{Concluding Remarks}

In this chapter, we have presented a neuro-symbolic reasoning framework for natural language processing. We view reasoning as intermediate thinking steps towards an end task, and we model it as discrete latent structures. Since discrete variables are not differentiable, we train them by reinforcement learning or its relaxation. In this way, we can provide explicit interpretation for the task. 

In future work, we plan to apply our neuro-symbolic framework to different NLP tasks and perform different types of reasoning, such as chunking~\cite{chunk} and logical reasoning~\cite{wu2021weakly}. Another important future direction is to automatically define the reasoning schema, which involves detecting important discrete latent structures to a task, as well as designing the neural architecture that can utilize the latent structures. Our long-term goal is to design an advanced neuro-symbolic framework that can detect semantic units, tell their relationships, and perform task-specific reasoning in an automated fashion.

\section*{Acknowledgments}

Lili Mou is supported in part by the Natural Sciences and Engineering Research Council of
Canada (NSERC) under grant No.~RGPIN2020-04465, the Amii Fellow Program, the Canada CIFAR AI Chair Program, a UAHJIC project, a donation from DeepMind, and the Digital Research Alliance of Canada (alliancecan.ca). Xianggen Liu acknowledges the support of the National Natural Science Foundation of China under Grant No.~62206192.

\bibliographystyle{plain}
\bibliography{reference.bib}

\end{document}